\pdfoutput=1

\documentclass[11pt]{article}

\usepackage[preprint]{acl}

\usepackage{times}
\usepackage{latexsym}
\usepackage{booktabs}
\usepackage{amsmath}
\usepackage{multirow}
\usepackage{amssymb}
\usepackage{subcaption}
\usepackage[most]{tcolorbox}
\usepackage{soul,xcolor} 
\sethlcolor{yellow}      

\usepackage[T1]{fontenc}

\usepackage[utf8]{inputenc}

\usepackage{microtype}

\usepackage{inconsolata}

\usepackage{subcaption}
\usepackage{graphicx}

\title{SEER: The Span-based Emotion Evidence Retrieval Benchmark}

\author{Aneesha Sampath \\ Computer Science \& Engineering \\  University of Michigan \\ \texttt{saneesha@umich.edu}
        \And  Oya Aran \\ Data Science \& AI \\ Procter \& Gamble \\ \texttt{aran.o@pg.com} \And
        Emily Mower Provost \\ Computer Science \& Engineering \\ University of Michigan \\ \texttt{emilykmp@umich.edu}}

\begin{document}
\maketitle
\begin{abstract}

We introduce the SEER (Span-based Emotion Evidence Retrieval) Benchmark to test Large Language Models’ (LLMs) ability to identify the specific spans of text that express emotion. Unlike traditional emotion recognition tasks that assign a single label to an entire sentence, SEER targets the underexplored task of \textit{emotion evidence} detection: pinpointing which exact phrases convey emotion. This span-level approach is crucial for applications like empathetic dialogue and clinical support, which need to know \textit{how} emotion is expressed, not just \textit{what} the emotion is. SEER includes two tasks: identifying emotion evidence within a single sentence, and identifying evidence across a short passage of five consecutive sentences. It contains new annotations for both emotion and emotion evidence on 1200 real-world sentences. We evaluate 14 open-source LLMs and find that, while some models approach average human performance on single-sentence inputs, their accuracy degrades in longer passages. Our error analysis reveals key failure modes, including overreliance on emotion keywords and false positives in neutral text.

\end{abstract}

\section{Introduction}

\begin{figure*}[ht]
    \centering
    \caption{SEER includes two tasks: single- and multi-sentence emotion evidence identification. Each has two prompt formats: Retrieve (extract exact spans) and Highlight (mark spans in context). Task objectives are identical across formats. The text is truncated in the figure for space, but not in actual LLM input/output.}
    \includegraphics[width=0.8\textwidth]{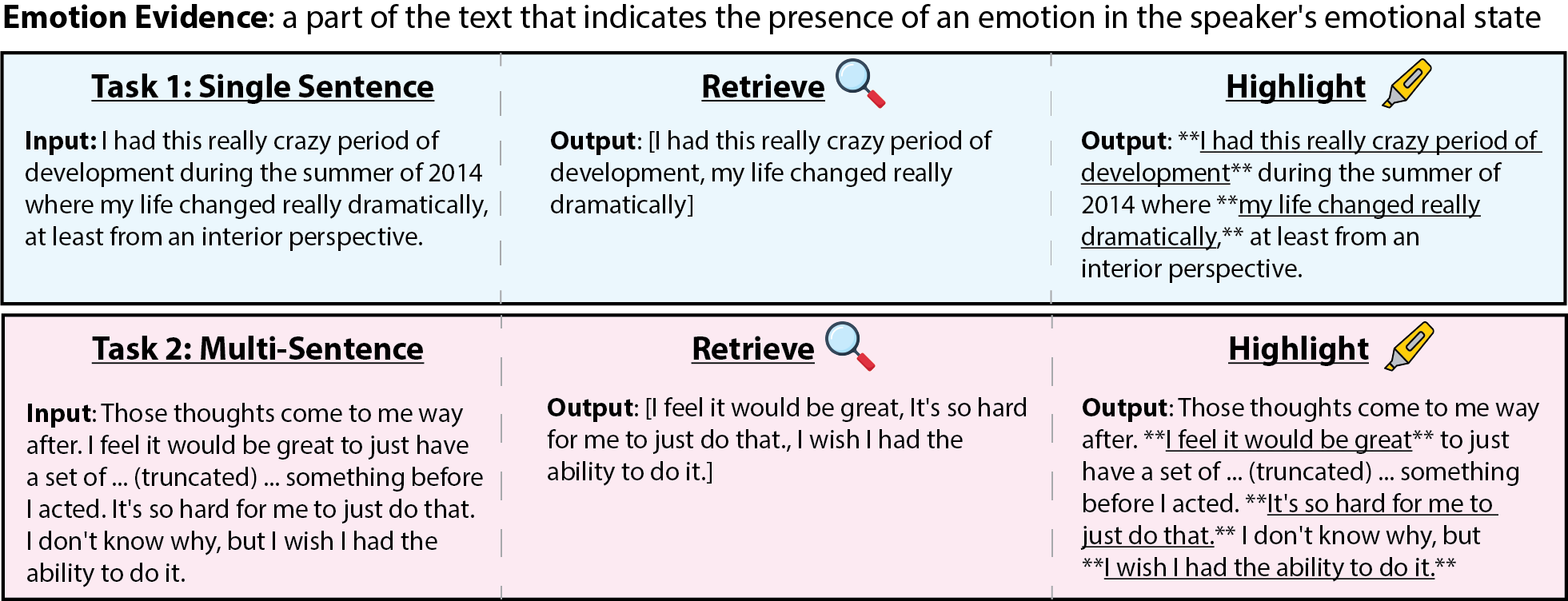}
    \label{fig:seer_overview}
\end{figure*}

We introduce the \underline{S}pan-based \underline{E}motion \underline{E}vidence \underline{R}etrieval (SEER) Benchmark, which evaluates Large Language Models (LLMs) on their ability to identify which spans of text express emotion in real-world discourse. SEER consists of two tasks: identifying emotion evidence within (1) a single sentence or (2) a short passage of five sentences. These span-level tasks differ from traditional emotion recognition benchmarks, which assign a single label to an entire utterance and do not isolate the exact phrases where emotion is expressed. SEER contains new annotations on 1200 real-world sentences, making it comparable in size to related emotion benchmarks \cite{sabour2024emobench}. For Task 1 (single sentence), we use GPT-4.1 \cite{achiam2023gpt} with human verification to label sentence-level emotion, with emotion evidence spans labeled by humans only. For Task 2 (five-sentence context), both the emotion labels and evidence spans are labeled by humans only. Figure~\ref{fig:seer_overview} illustrates the SEER benchmark.

\textit{Emotion evidence} refers to the spans of text that reveal a speaker's emotional state \cite{poria2021recognizing}. Identifying such spans is critical for applications like empathetic dialogue and clinical telehealth sessions, where responses depend not just on knowing \textit{what} emotion is present, but \textit{how} it is expressed linguistically. For example, knowing that a person is `sad' is less actionable than knowing that they said, `nobody cares anymore.'

Most prior work frames emotion recognition either as a sentence-level classification task \cite{bharti2022text, alvarez-gonzalez-etal-2021-uncovering-limits, wagner2023dawn} or as word-level tagging \cite{ito2020word, li2021word}. Sentence-level labels obscure which phrases convey emotion, while word-level tags often over-fragment emotionally coherent expressions \cite{hosseini2024disambiguating}. Span-level annotation offers a middle ground: localized enough for interpretability, but flexible enough to capture multi-word emotion cues.

Related work in speech emotion recognition focuses on identifying \textit{when} an emotion occurs in audio (e.g., at a certain timestamp), without linking those signals to the words used \cite{parthasarathy2016defining, aldeneh2017using}. As a result, these approaches cannot answer which parts of the linguistic content express emotion.

Progress on span-based emotion evidence detection has been limited by two main challenges: (1) a lack of datasets with span-level annotations (most provide only utterance-level labels \cite{busso2008iemocap, lotfian2017building}), and (2) a lack of datasets grounded in real-world discourse \cite{poria2019meld, busso2008iemocap}.

We evaluate 14 open-source LLMs on the SEER benchmark. Our results show that while several models approach average human performance in single-sentence settings, their accuracy declines in multi-sentence contexts. Key failure modes include fixation on explicit emotion keywords (e.g., `grateful') and false identification of emotion spans in neutral text. Future work could leverage SEER's span-level annotations to build models with better multi-word emotion identification and explore techniques to incorporate broader context to discourage keyword-matching. These directions can lead to LLMs that can more reliably pinpoint emotion expression in real-world discourse.

We publicly release all new annotations.\footnote{https://github.com/chailab-umich/SEER} Users must obtain access to the original datasets separately before working with the full SEER benchmark to comply with licensing requirements.

\section{Preliminaries}
\label{sec:preliminaries}

\subsection{Definition of Emotion}
\textit{Emotion} refers to a complex reaction involving experiential, behavioral, and physiological components, typically triggered by a personally meaningful event or situation \cite{apa_emotion}. Theories of emotion organize these within systematic frameworks.

The \textit{categorical emotion} theory posits that basic emotions developed in response to evolutionary needs. These emotions can include happiness, surprise, fear, sadness, anger, and disgust \cite{ekman1992argument}. The \textit{dimensional emotion} theory maps emotion along valence (negative to positive) and activation (calm to excited) \cite{harmon2017importance, russell1979affective}. Since text provides a stronger signal for valence compared to activation \cite{wagner2023dawn}, we focus on valence only for dimensional emotion. We conduct error analysis on the SEER tasks for categorical emotions and valence.

\subsection{Definition of Emotion Evidence}

\textit{Emotion evidence} is defined as ``a part of the text that indicates the presence of an emotion in the speaker's emotional state. It acts in the real world between the text and the reader" \cite{poria2021recognizing}. This should be distinguished from \textit{emotion cause}, which is the ``part of the text expressing the reason for the speaker to feel the emotion given by the emotion evidence" \cite{poria2021recognizing}.

\begin{tcolorbox}[
    colback=blue!1!white,
    colframe=yellow!50!black,
    title=\textsc{Example: Emotion Evidence vs Cause \cite{poria2021recognizing}},
    boxrule=0.4pt,
    arc=2pt,
    left=4pt,
    right=4pt,
    top=2pt,
    bottom=2pt,
    fonttitle=\scriptsize\bfseries,
    before upper={\setlength{\parskip}{2pt}\small},
]
\label{colorbox:ee_example}
\textsc{P\_A:} I have been accepted into graduate school! \\
\textsc{P\_B:} What an amazing accomplishment!

\vspace{0.7em}
\hrule
\vspace{0.3em}

\textsc{P\_B Label:} happy / positive

\vspace{0.5em}

\textsc{Cause:} accepted into graduate school \\
\textsc{Evidence:} amazing accomplishment
\end{tcolorbox}

\section{Related Work}
\subsection{Emotion Hotspot Detection}
Most emotion recognition work has targeted utterance-level classification, whereas \textit{emotion hotspots} identify specific points at which emotion shifts and intensifies \cite{huang2016detecting, huang2015investigation, parthasarathy2016defining}. Existing work on emotion hotspot detection has predominantly leveraged audio data. Some methods identify deviations from a baseline emotion state in valence-activation time-series traces \cite{parthasarathy2016defining, parthasarathy2018predicting}. Other approaches partition an audio stream to answer `which emotion appears when?' \cite{stemmer2023detection, wang2023speechzaion}. The output of these audio-based methods is a set of timestamps and corresponding emotion. They do not identify the specific spans used to express it. Our work addresses this complementary task: identifying the discrete, linguistic spans of emotion evidence directly from text, a capability that remains underexplored.

\subsection{Emotion Benchmarking in LLMs}
\begin{table}[t]
\tiny
\centering
\caption{Emotion benchmarks. H indicates hand-crafted, R indicates real-world, and S indicates LLM-generated.}
\begin{tabular}{l c c}
\toprule
\textbf{Benchmark} & \textbf{Focus} & \textbf{Data Type}  \\ 

\midrule
EmoBench \cite{sabour2024emobench} & scenario understanding & H \\ 
EmoLLMs \cite{liu2024emollms} & emotion recognition &  R \\ 
EmotionQueen \cite{liu2024emollms} & empathy generation & S \\ 
\textbf{SEER (ours)} & emotion evidence & R \\

\bottomrule
\end{tabular}
\label{tab:task_benchmark_comparison}
\end{table}

Benchmarks such as EmoBench, EmotionQueen, and EmoLLMs evaluate LLMs on emotion-related tasks but differ from SEER (see Table~\ref{tab:task_benchmark_comparison}). SEER evaluates the capability of LLMs to identify the precise spans where emotion evidence occurs. The other benchmarks have different goals, which we outline here. EmoBench \cite{sabour2024emobench} evaluates emotional reasoning through hand-crafted scenarios with multiple choice answers. Given an input, ``I have a teacher who gives the F grade as the highest mark... I saw he gave me an F," the LLM must identify the emotion of the speaker, and also the cause. This tests emotion recognition and emotion cause recognition, but uses hand-crafted scenarios. EmotionQueen \cite{chen2024emotionqueen} evaluates LLMs' ability to generate empathetic responses. Given a statement such as ``I’ve been busy with work all day," an empathetic model response might be ``Do you feel overwhelmed? Have you tried some ways to relax?", which provides emotional support, whereas a reply like ``Hard work!" is considered non-empathetic. EmoLLMs \cite{liu2024emollms} evaluates sentence-level emotion classification, where models assign emotion labels (e.g., happy, angry) to individual sentences. However, in all cases, these benchmarks do not localize the precise text that expresses emotion.

SEER tasks models with pinpointing the exact phrases that convey emotion, in both single-sentence and five-sentence passage settings. This focus on emotionally salient text spans grounded in real-world language fills a critical gap in existing emotion benchmarks.

\section{The SEER Benchmark}
The goal of SEER is to assess emotion evidence identification capabilities in LLMs. SEER comprises two primary tasks: single- and multi-sentence emotion evidence identification (Figure~\ref{fig:seer_overview}). All data are drawn from non-acted transcriptions of real-world speech (see Section~\ref{sec:datasets}) and annotated with both emotion labels and emotion evidence spans (see Sections~\ref{sec:task1_annotation} and~\ref{sec:task2_annotation} for annotation protocol).

\subsection{Task Versions: Retrieve and Highlight}
Each task has two versions: retrieve and highlight. In \textit{retrieve}, the LLM must output a series of spans. The output can be empty if the LLM identifies no spans of emotion evidence. In \textit{highlight}, the LLM must output the entire input passage, with the spans surrounded by `**' markers to indicate the start and end of an emotion evidence span.

These two prompt formats reflect real-world needs: Retrieve supports applications like evidence grounding or snippet retrieval, while Highlight supports scenarios requiring interpretable, in-context marking. To validate both formats, we conduct controlled prompting experiments with simple hand-crafted inputs (see Appendix~\ref{sec:prompting_appendix}). Success in these setups suggests that failures on SEER tasks stem from challenges in processing real-world emotion, not formatting or retrieval deficiencies.

\subsection{Task 1: Single-Sentence Emotion Evidence}
LLMs must identify all emotion evidence that occurs within a single, non-neutral sentence. The goal is to isolate short-form emotion expression.

\subsection{Task 2: Multi-Sentence Emotion Evidence}
LLMs must identify all emotion evidence that occurs within a series of five consecutive sentences. We select five sentences as a starting point. This length is manageable for annotation and analysis, yet long enough to see whether models can track emotion expression across coherent discourse. The goal is to test emotion evidence tracking in longer, more variable contexts, where the overall emotion may shift over the course of the passage.

\section{Datasets}
\label{sec:datasets}

\subsection{Pre-Existing Datasets}
We use samples from MSP-Podcast \cite{lotfian2017building} and MuSE \cite{jaiswal2019muse} for SEER. They contain non-acted speech (rather than scripted performances). We generate transcripts using Whisper\footnote{openai/whisper-large-v2} \cite{radford2023robust}.

MSP-Podcast contains non-acted English conversational speech from podcasts and includes both categorical and dimensional emotion annotations (version 1.11) \cite{lotfian2017building}. We use the subset of the data that overlaps with the MSP-Conversation corpus version 1.1 \cite{martinez2020msp}, which contains continuous time-series trace annotations of dimensional emotion on speech. This is to allow for future research combining the strengths of both continuous and sentence-level labels.  We use samples in the "Test1" split, totaling 2249 utterances.

The Multimodal Stressed Emotion (MuSE) dataset contains non-acted audiovisual English monologues \cite{jaiswal2019muse}. It includes crowdsourced annotations for dimensional emotion. It totals 2648 utterances.

We collect new text-based annotations for categorical emotion and valence to align with the LLMs’ input modality. The original MSP-Podcast and MuSE labels are audio- or video-based and the labels may not reflect textual cues. Using the original labels risks penalizing models for modality mismatch rather than genuine errors. In addition, MuSE lacks categorical emotion labels. Further, LLMs in Task 2 receive five-sentence context, which was not available to original annotators. Annotation details are in Sections~\ref{sec:task1_annotation} and~\ref{sec:task2_annotation}.

\subsection{Task 1 Data Annotation and Selection}
\label{sec:task1_annotation}
\paragraph{Single Sentence Filtering.} We filter the datasets to retain samples where utterances contain a single sentence only. We use NLTK \cite{bird2009natural} to tokenize by sentence. This leaves 1494 samples from MSP-Podcast and 1687 samples from MuSE.

\paragraph{GPT Labeling and Filtering.} We use GPT-4.1 \cite{achiam2023gpt} to ease the labeling burden. Prior work has shown GPT's capabilities for emotion labeling \cite{niu2024text, tarkka2024automated}. We complement it with human verification.

 We use the "gpt-4.1" model via the Azure OpenAI API to annotate the 1494 sentences from MSP-Podcast and 1687 sentences from MuSE for both valence (positive, negative, neutral) and categorical emotion (happy, sad, disgust, contempt, fear, angry, surprise, neutral). We select the eight categorical emotions that match the original label space from MSP-Podcast to encourage future research in the audio modality. The prompt is shown in Appendix \ref{sec:prompts}. We drop all samples that GPT-4.1 labeled as neutral. This leaves 488 sentences from MSP-Podcast and 854 sentences from MuSE. 

\begin{table}[ht]
\centering
\small
\caption{``\% Annotator" represents the \% of the time the annotators agreed with each other, and ``\% GPT" represents the \% of the time both annotators marked agree with the GPT label.}
\label{tab:gpt_agreement_table}
\begin{tabular}{lcc}
\toprule
\textbf{Dataset} & \textbf{\% Annotator} & \textbf{\% GPT} \\
\midrule
MuSE (Categorical) & 88.48\% & 87.31\% \\
MuSE (Valence)     & 90.72\% & 90.25\% \\
Podcast (Categorical) & 69.14\% & 60.29\% \\
Podcast (Valence)     & 71.05\% & 63.04\% \\
\bottomrule
\end{tabular}
\end{table}

\paragraph{Human Verification and Filtering.} Two trained student workers then independently indicated agreement or disagreement with the GPT-4.1 labels. This study is IRB-approved (HUM00273067). The annotator and GPT-4.1 agreement is in Table \ref{tab:gpt_agreement_table}.


We then filtered to only retain sentences where both annotators marked \textit{agree} for both the categorical and valence GPT-4.1 labels of a single sentence. This leaves 215 samples from MSP-Podcast and 703 samples from MuSE.

\paragraph{Emotion Class Balancing.} As a final filtering step, we balance the samples across the emotion classes by downsampling from over-represented classes. This leaves 30 samples per emotion class, except for surprise, which is slightly underrepresented with 20 samples, totaling 200 samples (103 from MSP-Podcast, 97 from MuSE). Since we balance by categorical emotion, valence is unbalanced since most of the emotion classes are negative. There are 155 negative and 45 positive samples. This is the final set of samples for Task 1.

\paragraph{Emotion Evidence Annotation.} The student workers received a short training to define emotion evidence and were instructed to openly discuss examples. In the final annotation step, they jointly identified and labeled the \textit{gold spans} of emotion evidence by discussing and highlighting the emotion evidence in the input text. This study is IRB-approved (HUM00273067).

\subsection{Task 2 Data Annotation and Selection}
\label{sec:task2_annotation}
Task 2 requires five \textit{consecutive} sentences in order to maintain semantic cohesion. We first split the original 2249 and 2648 utterances from MSP-Podcast and MuSE, respectively, into single sentences using NLTK. We retain instances with a series of five consecutive sentences. We then remove overlapping instances (i.e., only including sentences 4-8 and 9-13, instead of 4-8 and 5-9). This totals 200 sets of five consecutive sentences. 

The trained student workers were given each passage of five consecutive sentences, then discussed and jointly annotated the emotion (categorical and valence) of each sentence. They had access to all five sentences when annotating each sentence. This annotation step is necessary since the context of prior sentences can impact the emotion perception of a target sentence \cite{jaiswal2019muse}. This results in emotion labels for each sentence within the series of five sentences. We did not use GPT for multi-sentence emotion annotation, since it is not validated in prior work \cite{niu2024text}. 

For the final annotation step, the trained student workers jointly identified and labeled the gold spans of emotion evidence by discussing and highlighting their answers, as in Task 1.

\paragraph{Emotion Class Distribution}
Of the 1000 sentences (200 samples of 5 sentences each), they are 42.5\% neutral, 27.8\% happy, 11.6\% sad, 5.1\% surprise, 5.1\% fear, 4\% angry, 3.2\% contempt, and 0.7\% disgust. For valence, they are 42.7\% neutral, 30.7\% positive, and 26.6\% negative. We do not perform balancing due to the nature of the emotion shifts within passages of longer discourse. The most common emotion transition between adjacent sentences are maintaining the current emotion or transitioning to and from neutral. The exact distribution of emotion transitions is in Appendix \ref{sec:task2_emotion_transition_distribution}.

\section{Evaluation Metrics}
We use two primary evaluation metrics: token-level F1-score (F1) and cosine similarity (Sim). F1 is a common metric for span-extraction tasks \cite{rajpurkar2016squad}. It serves as a ``fuzzy-match" metric. In addition to F1, embedding similarity metrics have also emerged as a way to assess semantic similarity as opposed to exact matches \cite{Zhang2020BERTScore, arabzadeh2024adapting}, which can serve to reduce penalties for differences in span boundaries in SEER tasks. We use sentence-BERT \cite{reimers-gurevych-2019-sentence} to embed spans, and then compute the cosine similarity between the embeddings of the gold and predicted spans. We do not report exact-match accuracy due to the subjective nature of span boundaries. 

We use the Kuhn-Munkres Algorithm to align gold and predicted spans, which finds the optimal one-to-one matching between two sets \cite{luo2005coreference}. For example, consider two gold spans \( \{g_1, g_2\} \) and three predicted spans \( \{p_1, p_2, p_3\} \). The algorithm considers all possible matchings: $(g_1, p_1), (g_2, p_2)$, $(g_1, p_2), (g_2, p_1)$, etc. Each pairing is scored by ranking the similarity of the aligned span pairs, where similarity is defined as $\phi(g, p) = F_1(g, p)$, following \citet{luo2005coreference}. Since $g$ and $p$ are of unequal size, one span in $p$ remains unmatched.

We compute a modified score for both F1 and Sim that penalizes a model for predicting an incorrect number of spans. This approach ensures that a high score is achieved only when a model identifies the correct spans and the correct number of them. The score is calculated as the sum of the metrics from the aligned spans, normalized by the greater of the number of gold or predicted spans. This penalizes both irrelevant predicted spans (false positives) and missed gold spans (false negatives). The formula for metric $M$ (Sim or F1) is:

\begin{equation}
    M = \frac{\sum \text{metric}_{\text{matched\_spans}}}{\max(\#\text{GoldSpans}, \#\text{PredictedSpans})}
\end{equation}

\section{Implementation Details}

\subsection{Prompts}
We evaluate LLMs in two zero-shot prompt settings: Base prompting (Base) and chain-of-thought prompting (CoT), as in prior emotion benchmarking \cite{sabour2024emobench}. See Appendix \ref{sec:prompts}, Tables \ref{tab:system_prompts} and \ref{tab:SEER_user_prompts} for the exact prompts.

\subsection{LLMs Evaluated}
We evaluate the performance of 14 LLMs on the SEER benchmark. We select LLMs from the LLaMA \cite{dubey2024llama}, Qwen \cite{yang2025qwen3, yang2024qwen2}, Phi4 \cite{abdin2024phi, abdin2025phi}, and Gemma3 \cite{team2025gemma} families.

We select models that achieve F1 $\geq 0.5$ on prompting experiment three for further evaluation on the main SEER tasks. For the retrieve prompt, we retain all models with at least 1.7B parameters. For the highlight prompt, we retain models with at least 14B parameters, except for Qwen2.5-14B. 

\subsection{Experimental Setup}
We use the huggingface transformers library\footnote{https://github.com/huggingface/transformers} and load models in BF16. The full list of model checkpoint names is shown in Appendix Table \ref{tab:hf_checkpoints}. We use the default hyperparameters and allow a maximum of three retries. For each model, we report the average and standard deviation across five runs.

We run experiments on an HPC with NVIDIA A40 GPUs. We use one GPU for models in the 0.5-14B range, two for 32B, and four for 70-72B.

\section{Results}
\begin{table*}[t]
\centering
\tiny
\caption{F1 and cosine similarity (Sim) scores for Task 1 (Retrieve and Highlight). Each entry is averaged over five runs with standard deviations. `---' indicates the model was not evaluated in that prompt setting.}
\begin{tabular}{lcccccccc}
\toprule
\textbf{Model} 
& \multicolumn{2}{c}{\textbf{Retrieve (Base)}} 
& \multicolumn{2}{c}{\textbf{Retrieve (CoT)}}
& \multicolumn{2}{c}{\textbf{Highlight (Base)}} 
& \multicolumn{2}{c}{\textbf{Highlight (CoT)}} \\
\cmidrule(lr){2-3} \cmidrule(lr){4-5} \cmidrule(lr){6-7} \cmidrule(lr){8-9}
& F1 & Sim & F1 & Sim & F1 & Sim & F1 & Sim \\
\midrule

\multicolumn{9}{l}{\textbf{0.5--2B}} \\
\hspace{0.0em} Qwen 3 1.7B & 0.620 ± .004 & 0.638 ± .004 & 0.314 ± .011 & 0.361 ± .015 & --- & --- & --- & --- \\

\midrule
\multicolumn{9}{l}{\textbf{3--4B}} \\
\hspace{0.0em} LLaMA 3.2 3B & 0.193 ± .017 & 0.219 ± .018 & 0.213 ± .026 & 0.234 ± .021 & --- & --- & --- & --- \\
\hspace{0.0em} Phi 4 Mini 3.8B & 0.283 ± .013 & 0.285 ± .013 & 0.275 ± .010 & 0.289 ± .009 & --- & --- & --- & --- \\
\hspace{0.0em} Gemma 3 4B & 0.556 ± .002 & 0.583 ± .003 & 0.483 ± .019 & 0.502 ± .022 & --- & --- & --- & --- \\
\hspace{0.0em} Qwen 3 4B & 0.611 ± .004 & 0.632 ± .005 & 0.348 ± .006 & 0.389 ± .011 & --- & --- & --- & --- \\

\midrule
\multicolumn{9}{l}{\textbf{8B}} \\
\hspace{0.0em} LLaMA 3.1 8B & 0.487 ± .010 & 0.518 ± .006 & 0.410 ± .019 & 0.438 ± .021 & --- & --- & --- & --- \\
\hspace{0.0em} Qwen 3 8B & 0.646 ± .004 & 0.623 ± .002 & 0.487 ± .012 & 0.512 ± .009 & --- & --- & --- & --- \\

\midrule
\multicolumn{9}{l}{\textbf{14B}} \\
\hspace{0.0em} Phi 4 14B & 0.542 ± .009 & 0.567 ± .009 & 0.505 ± .013 & 0.521 ± .013 & 0.428 ± .008 & 0.488 ± .008 & 0.500 ± .007 & 0.534 ± .007 \\
\hspace{0.0em} Qwen 2.5 14B & 0.589 ± .003 & 0.607 ± .003 & 0.516 ± .005 & 0.531 ± .005 & --- & --- & --- & --- \\
\hspace{0.0em} Qwen 3 14B & 0.658 ± .004 & 0.677 ± .003 & 0.523 ± .024 & 0.548 ± .025 & 0.509 ± .009 & 0.561 ± .008 & 0.508 ± .013 & 0.558 ± .011 \\

\midrule
\multicolumn{9}{l}{\textbf{32B}} \\
\hspace{0.0em} Qwen 3 32B & \textbf{0.673 ± .006} & \textbf{0.693 ± .007} & 0.575 ± .014 & 0.596 ± .011 & 0.553 ± .013 & 0.578 ± .012 & 0.579 ± .009 & 0.606 ± .005 \\

\midrule
\multicolumn{9}{l}{\textbf{70--72B}} \\
\hspace{0.0em} LLaMA 3.1 70B & 0.437 ± .006 & 0.467 ± .006 & 0.501 ± .015 & 0.522 ± .011 & 0.300 ± .005 & 0.403 ± .007 & 0.392 ± .005 & 0.447 ± .008 \\
\hspace{0.0em} LLaMA 3.3 70B & 0.429 ± .008 & 0.469 ± .008 & 0.534 ± .009 & 0.564 ± .006 & 0.253 ± .007 & 0.359 ± .005 & 0.331 ± .012 & 0.407 ± .009 \\
\hspace{0.0em} Qwen 2.5 72B & 0.614 ± .006 & 0.633 ± .007 & 0.610 ± .012 & 0.625 ± .011 & 0.413 ± .005 & 0.483 ± .002 & 0.507 ± .006 & 0.533 ± .005 \\

\midrule
\multicolumn{9}{l}{\textbf{Human Annotator}} \\
\hspace{0.0em} Average & --- & --- & --- & --- & 0.458 ± .033 & 0.494 ± .034 & --- & --- \\
\hspace{0.0em} Best & --- & --- & --- & --- & 0.672 ± ----- & 0.675 ± ----- & --- & --- \\

\bottomrule
\end{tabular}
\label{tab:task1_scores}
\end{table*}

\begin{table*}[t]
\centering
\tiny
\caption{F1 and cosine similarity (Sim) scores for Task 2 (Retrieve and Highlight). Each entry is averaged over five runs with standard deviations. `---' indicates the model was not evaluated in that prompt setting.}
\begin{tabular}{lcccccccc}
\toprule
\textbf{Model} 
& \multicolumn{2}{c}{\textbf{Retrieve (Base)}} 
& \multicolumn{2}{c}{\textbf{Retrieve (CoT)}}
& \multicolumn{2}{c}{\textbf{Highlight (Base)}} 
& \multicolumn{2}{c}{\textbf{Highlight (CoT)}} \\
\cmidrule(lr){2-3} \cmidrule(lr){4-5} \cmidrule(lr){6-7} \cmidrule(lr){8-9}
& F1 & Sim & F1 & Sim & F1 & Sim & F1 & Sim \\
\midrule

\multicolumn{9}{l}{\textbf{0.5--2B}} \\
\hspace{0.0em} Qwen 3 1.7B & 0.253 ± .002 & 0.302 ± .003 & 0.248 ± .005 & 0.297 ± .007 & --- & --- & --- & --- \\

\midrule
\multicolumn{9}{l}{\textbf{3--4B}} \\
\hspace{0.0em} LLaMA 3.2 3B & 0.205 ± .009 & 0.234 ± .013 & 0.226 ± .007 & 0.255 ± .006 & --- & --- & --- & --- \\
\hspace{0.0em} Phi 4 Mini 3.8B & 0.224 ± .014 & 0.252 ± .013 & 0.238 ± .012 & 0.268 ± .012 & --- & --- & --- & --- \\
\hspace{0.0em} Gemma 3 4B & 0.368 ± .008 & 0.402 ± .008 & 0.332 ± .005 & 0.355 ± .007 & --- & --- & --- & --- \\
\hspace{0.0em} Qwen 3 4B & 0.335 ± .007 & 0.388 ± .006 & 0.281 ± .007 & 0.323 ± .005 & --- & --- & --- & --- \\

\midrule
\multicolumn{9}{l}{\textbf{8B}} \\
\hspace{0.0em} LLaMA 3.1 8B & 0.329 ± .006 & 0.367 ± .005 & 0.287 ± .009 & 0.326 ± .011 & --- & --- & --- & --- \\
\hspace{0.0em} Qwen 3 8B & 0.358 ± .006 & 0.362 ± .006 & 0.355 ± .008 & 0.379 ± .011 & --- & --- & --- & --- \\

\midrule
\multicolumn{9}{l}{\textbf{14B}} \\
\hspace{0.0em} Phi 4 14B & 0.342 ± .006 & 0.374 ± .009 & 0.357 ± .009 & 0.381 ± .009 & 0.259 ± .008 & 0.319 ± .007 & 0.346 ± .007 & 0.384 ± .008 \\
\hspace{0.0em} Qwen 2.5 14B & 0.362 ± .005 & 0.396 ± .005 & 0.348 ± .006 & 0.378 ± .006 & --- & --- & --- & --- \\
\hspace{0.0em} Qwen 3 14B & 0.406 ± .007 & 0.441 ± .006 & 0.354 ± .018 & 0.381 ± .020 & 0.263 ± .004 & 0.326 ± .005 & 0.315 ± .006 & 0.381 ± .009 \\

\midrule
\multicolumn{9}{l}{\textbf{32B}} \\
\hspace{0.0em} Qwen 3 32B & 0.405 ± .012 & 0.435 ± .012 & \textbf{0.410 ± .008} & \textbf{0.437 ± .008} & 0.322 ± .010 & 0.376 ± .009 & 0.388 ± .026 & 0.430 ± .026 \\

\midrule
\multicolumn{9}{l}{\textbf{70--72B}} \\
\hspace{0.0em} LLaMA 3.1 70B & 0.315 ± .006 & 0.352 ± .004 & 0.342 ± .021 & 0.374 ± .021 & 0.191 ± .006 & 0.268 ± .005 & 0.301 ± .013 & 0.365 ± .013 \\
\hspace{0.0em} LLaMA 3.3 70B & 0.280 ± .004 & 0.310 ± .003 & 0.345 ± .011 & 0.377 ± .013 & 0.150 ± .003 & 0.235 ± .003 & 0.247 ± .011 & 0.319 ± .014 \\
\hspace{0.0em} Qwen 2.5 72B & 0.391 ± .003 & 0.419 ± .004 & 0.350 ± .006 & 0.369 ± .006 & 0.284 ± .004 & 0.356 ± .004 & 0.363 ± .005 & 0.415 ± .009 \\

\midrule
\multicolumn{9}{l}{\textbf{Human Annotator}} \\
\hspace{0.0em} Average & --- & --- & --- & --- & 0.297 ± .069 & 0.336 ± .062 & --- & --- \\
\hspace{0.0em} Best & --- & --- & --- & --- & 0.506 ± ----- & 0.533 ± ----- & --- & --- \\

\bottomrule
\end{tabular}
\label{tab:task2_scores}
\end{table*}

\subsection{Task 1: Single-Sentence Emotion Evidence}
The results are shown in Table ~\ref{tab:task1_scores}. For the Retrieve-Base prompt, the Qwen-family models outperform all others, with every variant in the Qwen3 series achieving above 0.6 F1. Qwen3-32B achieves the highest score (0.673 F1, 0.693 Sim), while LLaMA3.2-3B performs the worst (0.193 F1). Notably, performance does not scale directly with model size: Qwen3-1.7B outperforms even the much larger LLaMA3.1/3.3-70B variants. Most models perform worse with Retrieve-CoT, with the exception of LLaMA3.2-3B, LLaMA3.1-70B, and LLaMA3.3-70B, which improve performance. We discuss CoT prompting in Section~\ref{sec:base_vs_cot_prompt}.

Performance drops for the Highlight prompt compared to Retrieve. This pattern is consistent with our prompting experiments, where we observe a typical performance drop of 0.3–0.4 F1 when comparing the same samples under Retrieve and Highlight settings (see Appendix~\ref{sec:prompting_appendix}). This drop is expected, as Highlight requires models to reproduce the input text verbatim with added markup. Any hallucination results in an automatic score of 0. However, models in the 14–32B range drop only about 0.15 F1, while the 70–72B models drop about 0.2 F1. All models tested with the Highlight prompt perform better or similarly under the CoT prompt than the Base prompt.

\subsection{Task 2: Multi-Sentence Emotion Evidence}
The results are in Table~\ref{tab:task2_scores}. Notably, no model exceeds 0.41 F1 in any prompt version on Task 2, underscoring the need for models capable of emotion evidence identification in extended passages.

The Qwen-family models again perform best on Retrieve-Base, with Qwen3-14B and Qwen3-32B leading with 0.406 and 0.405 F1, respectively. LLaMA3.2-3B remains the weakest model (0.205 F1), and the larger LLaMA3.1/3.3-70B variants are again outperformed by most smaller models. These results further reinforce that model size does not directly predict performance. 

Unlike Task 1 Retrieve, many models improve with CoT prompting in Task 2 Retrieve, including LLaMA3.2-3B, Phi4-Mini-3.8B, Phi4-14B, Qwen3-32B, LLaMA3.1-70B, and LLaMA3.3-70B. This suggest that CoT prompting is more effective for longer contexts, where reasoning steps may help localize relevant spans. For Highlight, this pattern persists: all models benefit from CoT prompts compared to Base. Highlight performance still falls short of Retrieve, as in Task 1.

\section{Human Performance Comparison}
\label{sec:human_performance_comparison}
We collected a crowdsourced human performance baseline for comparison with LLMs (IRB-approved). Details are in Appendix~\ref{sec:annotator_recruitment}. Each example received three independent annotations. The results are in Tables~\ref{tab:task1_scores} and~\ref{tab:task2_scores} in the `Human Annotator' rows. The `Average' row reflects the mean performance across all annotators, while the `Best' row reports the per-sample maximum: i.e., the annotation associated with the best-performing crowdsourced annotator, compared to the gold annotations, over each sample.

Many LLMs outperform Average, but only Qwen3-32B in Task 1 slightly exceeds the Best-Human. This suggests that at least one annotator often identifies the emotion evidence in the gold labels, but the annotations of crowdsourced annotators are of variable quality. It is expected that untrained workers underperform relative to expert annotators given the nuance of emotion evidence identification. The LLM-Best-Human gap is larger in Task 2 (about .1 F1) than in Task 1 (about .01 F1), indicating that a crowdsourced annotator outperforms LLMs in longer contexts.

\section{Error Analysis}

\subsection{Base and Chain-of-Thought Prompt}
\label{sec:base_vs_cot_prompt}

CoT prompting does not consistently improve performance across models in the Retrieve prompt setting (see Tables~\ref{tab:task1_scores} and~\ref{tab:task2_scores}). This aligns with findings from \citet{sabour2024emobench}, who reported that CoT prompting reduced or marginally changed performance on Emotion Intelligence tasks.

In Task 1 Retrieve, CoT prompting yields improvements for LLaMA3.1-70B and LLaMA3.3-70B, but degrades performance for all smaller models except LLaMA3.2-3B. In contrast, Task 2 Retrieve shows a less consistent trend: LLaMA3.2-3B, Phi4-Mini-3.8B, Qwen3-8B, Phi4-14B, Qwen3-32B, LLaMA3.1-70B, and LLaMA3.3-70B benefit from CoT prompting. This may reflect the nature of the longer input text in Task 2, where reasoning could assist span identification in longer passages.

\subsection{Hallucination Rates}
We define \textit{hallucination rate} as the fraction of predicted spans that do not appear in the original text. We normalize each predicted span by removing punctuation and converting to lowercase, and then compare it to the similarly-normalized transcription. We mark a span as `hallucinated' if it does not appear exactly as it is in the normalized text.

Models with lower hallucination scores reliably achieved higher F1 and similarity metrics. The Qwen family consistently showed the lowest hallucination rates across tasks and prompts.

Smaller models Phi4-mini-3.8B, LLaMA3.1-8B, and LLaMA3.2-3B exhibited the worst hallucination rates (16.3\%, 15.2\%, and 12.3\% for Task 1 Retrieve-Base). CoT had minimal impact on hallucination. These three models along with Qwen3-1.7B also exhibited high hallucination rates on Task 2 Retrieve-Base, contributing to their poor performance. The models evaluated in the Highlight prompts for both Task 1 and 2 consistently exhibit low hallucination rates ($\leq 5\%$).

\subsection{Errors by Emotion Category}

We discuss performance by emotion category on the Base prompts only. See Figures \ref{fig:task1_retrieve_errors}, \ref{fig:task1_highlight_errors}, \ref{fig:task2_retrieve_base_emotion_category_errors} and \ref{fig:task2_highlight_base_emotion_category_errors} in Appendix \ref{sec:emotion_category_errors} for visualizations. 

For Task 1, performance by emotion category is variable. In both Retrieve and Highlight, emotion evidence identification on sentences expressing \textit{disgust} and \textit{anger} perform best (Figures \ref{fig:task1_retrieve_categorical} and \ref{fig:task1_highlight_categorical}). The best performing model, Qwen3-32B, outperforms all other models on \textit{disgust} and \textit{happy} sentences in retrieve (Figure \ref{fig:task1_retrieve_categorical}), and outperforms other models on disgust, contempt, angry, and happy for highlight (Figure \ref{fig:task1_highlight_categorical}). The Qwen model family consistently performs better on negative sentences compared to positive sentences (Figures \ref{fig:task1_retrieve_valence} and \ref{fig:task1_highlight_valence}). This pattern is not consistent for the LLaMA, Gemma, and Phi families.

For Task 2, performance by emotion category resembles that of Task 1. However, unlike Task 1, Task 2 includes neutral sentences. The primary source of performance drop is the incorrect marking of emotion evidence in these neutral sentences. As shown in Figures~\ref{fig:task2_retrieve_nfpr} and~\ref{fig:task2_highlight_nfpr} in Appendix \ref{sec:emotion_category_errors}, models falsely identify emotion evidence in up to 50\% of neutral sentences. This high rate of neutral false positives degrades performance across models.

\subsection{Errors from Emotion Keyword Fixation}
We probe over-reliance on salient emotion words by checking if models extract emotion keywords in isolation rather than the full span. We use the Empath lexicon ~\cite{fast2016empath} to identify instances where transcripts contain terms from the \textit{positive-emotion} and \textit{negative-emotion} categories. In 61 of the 200 sentences in Task 1 and 84 of the 1000 sentences in Task 2, a gold span contains an emotion keyword. We define a \textit{fixation} as any prediction in which the model identifies only the keyword itself (e.g., predicting `disgust' when the gold span is `I would like to state my utter disgust.').

For both tasks, this error pattern appears most prominently in Highlight. For Highlight-Base, LLaMA3.1-70B and LLaMA3.3-70B exhibit high fixation rates of 27.9\% and 26.6\% for Task 1, and 36.2\% and 53.6\% for Task 2, respectively, where samples with an emotion keyword default to isolated words despite the emotion expressions themselves consisting of longer spans. In contrast, Qwen3-32B, the best-performing model, has only 1.6\% of samples with this behavior in Task 1 and 5.7\% in Task 2 (lowest fixation rate of all models). For Retrieve, most models have 0\% fixation rates in Task 1, with the exception of LLaMA3.2-3B (3.6\%), Qwen3-4B (1\%), Phi4 (0.3\%), LLaMA3.1-70B (1.6\%), and LLaMA3.3-70B (6.9\%). Similarly, in Task 2, the highest fixation rate is 7.1\% (LLaMA3.2-3B), with most models falling within 0-3\% fixation (except for Qwen3-4B with 5.7\%).

CoT prompting partially mitigates this behavior for larger models. Fixation rates for LLaMA3.1-70B and LLaMA3.3-70B drop to 15.4\% and 16.1\%, for Task 1 Highlight-CoT and to 14.3\% and 36.2\% for Task 2, respectively, suggesting that reasoning steps can encourage more holistic span identification. However, this trend does not generalize across scales. In Task 1 Retrieve-CoT, smaller models Qwen3-1.7B and Qwen3-4B show increased fixation under CoT (11.8\% and 9.8\%, respectively), despite minimal errors with the base prompt.

These results underscore that keyword fixation is a nuanced failure mode. While CoT can guide larger models toward more nuanced span identification, it may also backfire in smaller models by drawing attention to more obvious word-level cues. Crowdsourced annotators also exhibit about 3\% fixation rate in both tasks, suggesting that even human annotators are prone to this error mode.

\section{Conclusion}
In this paper, we propose the SEER Benchmark for evaluating LLM capability in emotion evidence identification. SEER comprises two tasks: single-sentence and multi-sentence emotion evidence identification in real-world discourse. We collect new annotations for 1200 sentences for emotion category, valence, and evidence. We evaluate SEER on 14 open-source LLMs and conduct a comprehensive error analysis. We find that models can somewhat reliably identify emotion evidence in single sentences, however, these models falsely identify emotion evidence in neutral sentences in multi-sentence contexts. Key error modes also include fixation on emotion keywords and modification of the input text (hallucination). Of the models we evaluate, Qwen3-32B performs the best in both SEER tasks.

Future work may explore whether performance on SEER aligns with standard emotion classification by evaluating the same LLMs on related tasks. SEER can also be evaluated on closed-source and reasoning models. Finally, SEER could be adapted to the audio modality by using the original datasets' annotations for evaluation on audio-LLMs.

\section*{Limitations}
\textbf{Sample Size.} SEER is limited to 200 samples in Task 1 and 200 samples in Task 2. While this contains high-quality annotations for emotion evidence, emotion valence, and emotion category, and also is similar to the size of other emotion benchmarks \cite{sabour2024emobench}, we acknowledge that our dataset scale is limited, and could benefit from additional samples. 

\textbf{Prompt Tuning.} We acknowledge that LLM outputs are highly sensitive to input prompts and that additional techniques could influence performance. We conducted extensive prompting experiments to mitigate this effect. Prompt design adjustments may impact the exact numerical scores, however we argue that they are unlikely to alter the overall trends observed across the tasks, as observed when comparing base and chain-of-thought prompting results.

\section*{Acknowledgments}
This work was supported by Procter \& Gamble. We used ChatGPT to assist with minor language editing in the paper and minor code debugging.

We thank Yara El-Tawil and James Tavernor for assisting with data annotation. We thank Stuart Synakowski, Minxue Niu, Mimansa Jaiswal, Arian Raje, and Victoria Lin for providing project feedback.

\bibliography{custom}

\appendix

\section{Hand-crafted Sentences}
\label{sec:handcrafted_appendix}
We design a set of hand-crafted sentences for the prompting experiments (detailed in Appendix~\ref{sec:prompting_appendix}). The sentences include both neutral and emotion expressions. This allows us to evaluate whether models can reliably execute instruction-following behavior and return outputs in the expected format without the challenges of subtle and potentially ambiguous real-world language. Demonstrating robust performance under these conditions ensures that any failures observed in the main SEER tasks, which use real-world discourse, are not due to fundamental retrieval limitations or format misalignment, but instead reflect genuine challenges in understanding real-world emotion expression.

The sentences contain no repeating bi-grams across both the neutral and emotion sentences. The reasoning for this is described in Appendix \ref{sec:appendix_sentence_constraints}. We construct ten neutral sentences, two sentences for each categorical emotion in MSP-Podcast \cite{lotfian2017building} (happy, sad, disgust, contempt, fear, anger, surprise), and five sentences for each valence (positive, negative). The full list of sentences is shown in Table \ref{tab:appendix_handcrafted_sentences}.

\begin{table*}[t]
\centering
\small
\caption{Emotion-labeled sentences used in prompting experiments.}
\begin{tabular}{llr}
\toprule
\textbf{Emotion} & \textbf{Sentence} & \textbf{Gold Spans} \\
\midrule
\multirow{2}{*}{happy} 
& Everything felt perfect this morning. & Everything felt perfect \\
& This evening feels like a beautiful dream. & a beautiful dream \\
\midrule
\multirow{2}{*}{sad} 
& Tears filled my eyes uncontrollably. & Tears filled my eyes uncontrollably. \\
& I felt so empty inside. & I felt so empty inside. \\
\midrule
\multirow{2}{*}{disgust} 
& The sight turned my stomach upside down. & turned my stomach upside down \\
& I shivered because it was so revolting. & I shivered / it was so revolting \\
\midrule
\multirow{2}{*}{contempt} 
& I sneered at the pathetic excuse. & I sneered / pathetic excuse \\
& I discarded his nonsense as laughable. & I discarded his nonsense as laughable. \\
\midrule
\multirow{2}{*}{fear} 
& The door creak sent chills down my spine. & chills down my spine \\
& I began trembling when I heard the thunder. & trembling \\
\midrule
\multirow{2}{*}{anger} 
& My blood is boiling with rage. & boiling with rage \\
& The disrespectful comment made me feel like I was going to explode. & I was going to explode \\
\midrule
\multirow{2}{*}{surprise} 
& I blinked in disbelief. & I blinked in disbelief. \\
& My jaw dropped seeing the unexpected. & My jaw dropped seeing the unexpected. \\
\midrule
\multirow{5}{*}{positive} 
& Warm laughter filled the room. & Warm laughter \\
& The sunlight was warm and inviting. & warm and inviting \\
& Aromas of fresh flowers brought a smile to my face. & brought a smile to my face \\
& Reflection on past events gives me hope for the future. & gives me hope \\
& I am grateful for all support I received. & I am grateful \\
\midrule
\multirow{5}{*}{negative} 
& My hopes crumbled upon hearing the truth. & My hopes crumbled \\
& The regret of my actions haunted me. & regret / haunted me \\
& The tension was unbearable. & The tension was unbearable. \\
& I did not appreciate the comments. & I did not appreciate \\
& Dread filled me as I thought about the consequences. & Dread filled me \\
\midrule 
\multirow{10}{*}{neutral} 
& A book lies on the desk. \\
& A clock shows the time. \\
& Light travels in straight lines. \\
& Clouds exist in the sky. \\
& Rocks form over long periods. \\
& Pens leave ink marks on paper. \\
& Windows reflect ambient light. \\
& Books contain printed pages. \\
& A key is used to unlock a door. \\
& Soil is made of rocks and minerals. \\
\bottomrule
\end{tabular}
\label{tab:appendix_handcrafted_sentences}
\end{table*}

\section{Prompting Experiments}
\label{sec:prompting_appendix}

In this section, we detail our experiments for the \textit{retrieval} and \textit{highlight} prompting styles. These allow us to identify which LLMs are capable of producing responses in our desired format.

\subsection{Motivation}
In some applications, it may be sufficient to have models retrieve only exact quotes of emotion evidence. In others, it may be necessary to place the emotion evidence back within the original context in which it was communicated. We observe that smaller models are generally capable of reproducing given text without hallucination, however they fail to reliably "highlight" a sentence within the given text. In order properly to evaluate emotion evidence capabilities in both smaller and larger models, we develop two sets of prompts: \textit{retrieval} and \textit{highlight}. The \textit{retrieval} prompts evaluate LLM capability in retrieving exact quotes of emotion evidence only. The \textit{highlight} prompts evaluate LLM capability in highlighting the exact regions of emotion evidence while also retrieving the original text without hallucination. These experiments are summarized in Table \ref{tab:prompting_experiment_overview}.

The goal of these experiments is to assess retrieval capacity, not to assess ability to identify subtle emotion evidence. Thus, for the prompting experiments, we use hand-crafted data only. These sentences are designed to unambiguously express either neutrality or a categorical emotion. See Appendix \ref{sec:handcrafted_appendix} for the full list of sentences. 

\subsection{Experiments}
We include four experiments. Experiment 0 is a baseline experiment to assess LLM capability in reproducing passages without any modification. The LLM must reproduce the exact input text, unmodified. Experiment 1 requires the LLM to identify one span of text that occurs within the original text. The target span is provided in the prompt. This experiment uses neutral sentences only. Experiment 2 requires the LLM to identify all spans of emotion evidence in the text. The LLM is given three span options in the prompt itself, in which either one or two of the three spans are correct. Experiment 3 requires the LLM to identify all spans of emotion evidence in the text, without any options provided in the prompt.

Each experiment has a retrieval and highlight version, except for Experiment 0, since there is no span identification involved. In the retrieval versions, the LLM must retrieve the specific spans of text only. In the highlight versions, the LLM must return the entire input text, and mark the specific spans by surrounding the spans with `**' markers.

We set the number of sentences (n\_sentences) that the LLM must retrieve in $[1, 10]$, to identify if errors stem from the length of the input/output or from the nature of the experiment. For Experiment 0, we create ten variants for each $n \in \text{n\_sentences}$, where $n$ neutral sentences are randomly samples and randomly shuffled. This totals 100 samples. For Experiment 1, we use the same logic as Experiment 0, except with $\text{n\_sentences} \in [2,10]$, since we need at least two sentences in order to be able to identify the target sentence. This totals 90 samples. For Experiment 2 and 3, which contain the same samples, we also set $\text{n\_sentences} \in [2,10]$. We create two variants for each emotion sentence, where three sets of neutral sentences are randomly selected and shuffled. The emotion sentence is randomly placed within the neutral sentences. This totals 432 samples.

\begin{table*}[t]
\centering
\small
\caption{Overview of prompting experiments. Each experiment tests a different task objective. Retrieval variants require the LLM to output the relevant span. Highlight variants require the LLM to reproduce the input with the relevant span marked using a delimiter (\texttt{**...**}).}
\begin{tabular}{l p{10cm}}
\toprule
\textbf{Experiment} & \textbf{Task Description} \\
\midrule
Exp 0 & Reproduce the input text exactly, with no modifications. \\
Exp 1 & Identify one span given the exact span in the instructions. \\
Exp 2 & Identify all spans of emotion evidence given three span options in the instructions. \\
Exp 3 & Identify all spans of emotion evidence in the original text. \\
\bottomrule
\end{tabular}
\label{tab:prompting_experiment_overview}
\end{table*}

\subsection{Metrics and Constraints}
\label{sec:appendix_sentence_constraints}
We use metrics Exact-match accuracy (EM) and token-level F1-Score (F1) to evaluate LLM performance on the prompting experiments. These metrics are used in the SQuAD benchmark for question-answering \cite{rajpurkar2016squad}. We use these to compare the LLM output to the expected output.


\subsection{Results}
\begin{table}[t]
\centering
\small
\caption{Exact-match accuracy (EM) and F1 scores for prompting Experiment 0 (baseline reproduction). We report the average and standard deviation over five runs.}
\begin{tabular}{l c c}
\toprule
\textbf{Model} & EM & F1 \\
\midrule

\multicolumn{3}{l}{\textbf{0.5--2B}} \\
\hspace{0.0em} Qwen 3 0.6B & 1.000 ± .000 & 1.000 ± .000 \\
\hspace{0.0em} Gemma 3 1B & 1.000 ± .000 & 1.000 ± .000 \\
\hspace{0.0em} LLaMA 3.2 1B & 0.730 ± .012 & 0.962 ± .011 \\
\hspace{0.0em} Qwen 3 1.7B & 0.832 ± .013 & 0.943 ± .005 \\

\multicolumn{3}{l}{\textbf{3--4B}} \\
\hspace{0.0em} LLaMA 3.2 3B & 0.990 ± .000 & 1.000 ± .000 \\
\hspace{0.0em} Phi 4 Mini 3.8B & 0.710 ± .019 & 0.956 ± .003 \\
\hspace{0.0em} Gemma 3 4B & 1.000 ± .000 & 1.000 ± .000 \\
\hspace{0.0em} Qwen 3 4B & 1.000 ± .000 & 1.000 ± .000 \\

\midrule
\multicolumn{3}{l}{\textbf{8B}} \\
\hspace{0.0em} LLaMA 3.1 8B & 1.000 ± .000 & 1.000 ± .000 \\
\hspace{0.0em} Qwen 3 8B & 1.000 ± .000 & 1.000 ± .000 \\

\midrule
\multicolumn{3}{l}{\textbf{14B}} \\
\hspace{0.0em} Phi 4 14B & 0.986 ± .015 & 0.998 ± .002 \\
\hspace{0.0em} Qwen 2.5 14B & 1.000 ± .000 & 1.000 ± .000 \\
\hspace{0.0em} Qwen 3 14B & 1.000 ± .000 & 1.000 ± .000 \\

\midrule
\multicolumn{3}{l}{\textbf{32B}} \\
\hspace{0.0em} Qwen 3 32B & 1.000 ± .000 & 1.000 ± .000 \\

\midrule
\multicolumn{3}{l}{\textbf{70--72B}} \\
\hspace{0.0em} LLaMA 3.1 70B & 1.000 ± .000 & 1.000 ± .000 \\
\hspace{0.0em} LLaMA 3.3 70B & 1.000 ± .000 & 1.000 ± .000 \\
\hspace{0.0em} Qwen 2.5 72B & 1.000 ± .000 & 1.000 ± .000 \\

\bottomrule
\end{tabular}
\label{tab:prompting_exp0}
\end{table}

\begin{table*}[t]
\centering
\tiny
\caption{Token-level F1 and cosine similarity (Sim) scores for retrieval prompting experiments. We report the average and standard deviation over five runs. \textit{Categorical} and \textit{Valence} indicate the model was evaluated on sentences crafted to target either a specfic emotion category or emotion valence.}
\begin{tabular}{l c c c c c c c c c c}
\toprule
\textbf{Model} 
& \multicolumn{2}{c}{\textbf{Exp 1}} 
& \multicolumn{2}{c}{\textbf{Exp 2 – Categorical}} 
& \multicolumn{2}{c}{\textbf{Exp 2 – Valence}} 
& \multicolumn{2}{c}{\textbf{Exp 3 – Categorical}} 
& \multicolumn{2}{c}{\textbf{Exp 3 – Valence}} \\
\cmidrule(lr){2-3} \cmidrule(lr){4-5} \cmidrule(lr){6-7} \cmidrule(lr){8-9} \cmidrule(lr){10-11}
& Sim & F1 & Sim & F1 & Sim & F1 & Sim & F1 & Sim & F1 \\
\midrule

\multicolumn{11}{l}{\textbf{0.5–2B}} \\
\hspace{0.0em} Qwen 3 0.6B & 0.814 ± .004 & 0.844 ± .003 & 0.430 ± .007 & 0.461 ± .008 & 0.382 ± .005 & 0.410 ± .003 & 0.438 ± .009 & 0.404 ± .013 & 0.314 ± .005 & 0.279 ± .014 \\
\hspace{0.0em} Gemma 3 1B & 0.768 ± .004 & 0.778 ± .002 & 0.594 ± .004 & 0.541 ± .005 & 0.578 ± .002 & 0.562 ± .003 & 0.439 ± .005 & 0.567 ± .006 & 0.419 ± .007 & 0.492 ± .008 \\
\hspace{0.0em} LLaMA 3.2 1B & 0.436 ± .008 & 0.192 ± .005 & 0.350 ± .006 & 0.343 ± .005 & 0.357 ± .013 & 0.364 ± .010 & 0.167 ± .012 & 0.064 ± .006 & 0.168 ± .012 & 0.074 ± .015 \\
\hspace{0.0em} Qwen 3 1.7B & 0.824 ± .002 & 0.833 ± .001 & 0.683 ± .005 & 0.689 ± .004 & 0.582 ± .005 & 0.595 ± .004 & 0.858 ± .005 & 0.915 ± .003 & 0.843 ± .006 & 0.885 ± .006 \\

\midrule
\multicolumn{11}{l}{\textbf{3–4B}} \\
\hspace{0.0em} LLaMA 3.2 3B & 0.817 ± .014 & 0.782 ± .011 & 0.700 ± .004 & 0.711 ± .004 & 0.548 ± .008 & 0.586 ± .006 & 0.824 ± .012 & 0.810 ± .008 & 0.845 ± .011 & 0.818 ± .013 \\
\hspace{0.0em} Phi 4 Mini 3.8B & 0.879 ± .007 & 0.877 ± .005 & 0.722 ± .005 & 0.708 ± .011 & 0.598 ± .009 & 0.578 ± .014 & 0.678 ± .015 & 0.815 ± .007 & 0.594 ± .015 & 0.777 ± .010 \\
\hspace{0.0em} Gemma 3 4B & 0.877 ± .001 & 0.883 ± .001 & 0.567 ± .001 & 0.601 ± .002 & 0.488 ± .003 & 0.547 ± .003 & 0.858 ± .003 & 0.831 ± .005 & 0.789 ± .005 & 0.780 ± .006 \\
\hspace{0.0em} Qwen 3 4B & 0.876 ± .000 & 0.891 ± .001 & 0.725 ± .005 & 0.725 ± .004 & 0.564 ± .002 & 0.575 ± .002 & 0.949 ± .005 & 0.962 ± .004 & 0.953 ± .005 & 0.973 ± .004 \\

\midrule
\multicolumn{11}{l}{\textbf{8B}} \\
\hspace{0.0em} LLaMA 3.1 8B & 0.876 ± .006 & 0.890 ± .004 & 0.754 ± .003 & 0.754 ± .006 & 0.642 ± .009 & 0.626 ± .006 & 0.979 ± .003 & 0.990 ± .002 & 0.982 ± .004 & 0.999 ± .001 \\
\hspace{0.0em} Qwen 3 8B & 0.833 ± .002 & 0.870 ± .001 & 0.699 ± .003 & 0.722 ± .003 & 0.614 ± .002 & 0.634 ± .002 & 0.946 ± .002 & 0.998 ± .001 & 0.943 ± .006 & 0.993 ± .003 \\

\midrule
\multicolumn{11}{l}{\textbf{14B}} \\
\hspace{0.0em} Phi 4 14B & 0.886 ± .006 & 0.884 ± .012 & 0.781 ± .008 & 0.798 ± .007 & 0.734 ± .005 & 0.755 ± .004 & 0.986 ± .006 & 0.997 ± .001 & 0.995 ± .006 & 0.999 ± .002 \\
\hspace{0.0em} Qwen 2.5 14B & 0.839 ± .001 & 0.879 ± .001 & 0.731 ± .002 & 0.726 ± .001 & 0.631 ± .004 & 0.670 ± .004 & 0.993 ± .002 & 0.997 ± .000 & 0.991 ± .000 & 0.989 ± .000 \\
\hspace{0.0em} Qwen 3 14B & 0.865 ± .000 & 0.912 ± .001 & 0.677 ± .003 & 0.687 ± .004 & 0.641 ± .003 & 0.675 ± .004 & 1.000 ± .000 & 1.000 ± .000 & 1.000 ± .000 & 1.000 ± .000 \\

\midrule
\multicolumn{11}{l}{\textbf{32B}} \\
\hspace{0.0em} Qwen 3 32B & 0.875 ± .003 & 0.899 ± .001 & 0.740 ± .004 & 0.739 ± .005 & 0.698 ± .005 & 0.740 ± .003 & 1.000 ± .000 & 1.000 ± .000 & 1.000 ± .000 & 1.000 ± .000 \\

\midrule
\multicolumn{11}{l}{\textbf{70–72B}} \\
\hspace{0.0em} LLaMA 3.1 70B & 0.828 ± .005 & 0.858 ± .003 & 0.777 ± .003 & 0.763 ± .002 & 0.677 ± .003 & 0.679 ± .003 & 0.937 ± .004 & 0.946 ± .004 & 0.928 ± .004 & 0.920 ± .003 \\
\hspace{0.0em} LLaMA 3.3 70B & 0.838 ± .002 & 0.861 ± .002 & 0.821 ± .004 & 0.802 ± .005 & 0.713 ± .003 & 0.704 ± .001 & 0.900 ± .003 & 0.918 ± .002 & 0.922 ± .001 & 0.942 ± .002 \\
\hspace{0.0em} Qwen 2.5 72B & 0.834 ± .002 & 0.871 ± .002 & 0.846 ± .002 & 0.832 ± .002 & 0.791 ± .002 & 0.816 ± .003 & 0.999 ± .001 & 0.999 ± .001 & 0.996 ± .000 & 0.997 ± .000 \\

\bottomrule
\end{tabular}
\label{tab:retrieval_prompting_exp_tab}
\end{table*}

\begin{table*}[t]
\centering
\tiny
\caption{F1 and similarity (Sim) scores for highlight prompting experiments (Exp 1, Exp 2 – Categorical, Exp 2 – Valence, and Exp 3). We report the average and standard deviation over five runs. \textit{Categorical} and \textit{Valence} indicate the model was evaluated on sentences crafted to target either a specfic emotion category or emotion valence.}
\begin{tabular}{l cc cc cc cc cc}
\toprule
\textbf{Model} 
& \multicolumn{2}{c}{\textbf{Exp 1}} 
& \multicolumn{2}{c}{\textbf{Exp 2 – Categorical}} 
& \multicolumn{2}{c}{\textbf{Exp 2 – Valence}} 
& \multicolumn{2}{c}{\textbf{Exp 3 – Categorical}} 
& \multicolumn{2}{c}{\textbf{Exp 3 – Valence}} \\
\cmidrule(lr){2-3} \cmidrule(lr){4-5} \cmidrule(lr){6-7} \cmidrule(lr){8-9} \cmidrule(lr){10-11}
& F1 & Sim & F1 & Sim & F1 & Sim & F1 & Sim & F1 & Sim \\
\midrule

\multicolumn{11}{l}{\textbf{0.5--2B}} \\
\hspace{0.0em} Qwen 3 0.6B & 0.028 ± .004 & 0.046 ± .005 & 0.139 ± .007 & 0.183 ± .005 & 0.118 ± .007 & 0.117 ± .006 & 0.085 ± .005 & 0.122 ± .004 & 0.086 ± .005 & 0.086 ± .003 \\
\hspace{0.0em} Gemma 3 1B & 0.172 ± .008 & 0.245 ± .006 & 0.078 ± .002 & 0.098 ± .004 & 0.073 ± .002 & 0.075 ± .002 & 0.023 ± .001 & 0.029 ± .002 & 0.029 ± .000 & 0.035 ± .001 \\
\hspace{0.0em} LLaMA 3.2 1B & 0.000 ± .000 & 0.000 ± .000 & 0.024 ± .003 & 0.026 ± .004 & 0.026 ± .006 & 0.028 ± .005 & 0.039 ± .004 & 0.044 ± .004 & 0.036 ± .004 & 0.036 ± .004 \\
\hspace{0.0em} Qwen 3 1.7B & 0.659 ± .004 & 0.681 ± .004 & 0.471 ± .003 & 0.522 ± .003 & 0.371 ± .003 & 0.399 ± .003 & 0.287 ± .003 & 0.367 ± .003 & 0.259 ± .003 & 0.294 ± .003 \\

\multicolumn{11}{l}{\textbf{3--4B}} \\
\hspace{0.0em} LLaMA 3.2 3B & 0.104 ± .011 & 0.117 ± .016 & 0.202 ± .010 & 0.218 ± .010 & 0.175 ± .012 & 0.179 ± .011 & 0.048 ± .009 & 0.056 ± .011 & 0.056 ± .008 & 0.061 ± .011 \\
\hspace{0.0em} Phi 4 Mini 3.8B & 0.035 ± .007 & 0.038 ± .009 & 0.348 ± .023 & 0.373 ± .025 & 0.397 ± .027 & 0.414 ± .029 & 0.329 ± .011 & 0.380 ± .013 & 0.322 ± .022 & 0.360 ± .020 \\
\hspace{0.0em} Gemma 3 4B & 0.264 ± .004 & 0.247 ± .005 & 0.283 ± .003 & 0.304 ± .003 & 0.247 ± .008 & 0.251 ± .008 & 0.227 ± .006 & 0.252 ± .007 & 0.180 ± .005 & 0.197 ± .005 \\
\hspace{0.0em} Qwen 3 4B & 0.562 ± .010 & 0.592 ± .005 & 0.665 ± .004 & 0.693 ± .003 & 0.651 ± .007 & 0.657 ± .007 & 0.207 ± .004 & 0.229 ± .004 & 0.267 ± .004 & 0.293 ± .002 \\

\midrule
\multicolumn{11}{l}{\textbf{8B}} \\
\hspace{0.0em} LLaMA 3.1 8B & 0.077 ± .009 & 0.095 ± .008 & 0.049 ± .004 & 0.054 ± .004 & 0.053 ± .002 & 0.055 ± .004 & 0.078 ± .004 & 0.085 ± .005 & 0.084 ± .006 & 0.101 ± .006 \\
\hspace{0.0em} Qwen 3 8B & 0.667 ± .009 & 0.674 ± .008 & 0.263 ± .006 & 0.289 ± .006 & 0.370 ± .006 & 0.366 ± .005 & 0.110 ± .006 & 0.129 ± .006 & 0.148 ± .003 & 0.158 ± .003 \\

\midrule
\multicolumn{11}{l}{\textbf{14B}} \\
\hspace{0.0em} Phi 4 14B & 0.906 ± .018 & 0.895 ± .020 & 0.808 ± .004 & 0.825 ± .003 & 0.714 ± .012 & 0.721 ± .014 & 0.698 ± .006 & 0.761 ± .003 & 0.606 ± .006 & 0.653 ± .003 \\
\hspace{0.0em} Qwen 2.5 14B & 0.714 ± .013 & 0.718 ± .009 & 0.761 ± .005 & 0.777 ± .004 & 0.706 ± .005 & 0.725 ± .005 & 0.484 ± .005 & 0.536 ± .005 & 0.366 ± .005 & 0.448 ± .004 \\
\hspace{0.0em} Qwen 3 14B & 0.877 ± .005 & 0.878 ± .007 & 0.695 ± .004 & 0.711 ± .004 & 0.608 ± .003 & 0.616 ± .006 & 0.620 ± .003 & 0.679 ± .004 & 0.499 ± .005 & 0.548 ± .005 \\

\midrule
\multicolumn{11}{l}{\textbf{32B}} \\
\hspace{0.0em} Qwen 3 32B & 0.975 ± .004 & 0.961 ± .002 & 0.738 ± .005 & 0.762 ± .004 & 0.683 ± .004 & 0.685 ± .003 & 0.706 ± .002 & 0.773 ± .002 & 0.597 ± .004 & 0.649 ± .002 \\

\midrule
\multicolumn{11}{l}{\textbf{70--72B}} \\
\hspace{0.0em} LLaMA 3.1 70B & 0.850 ± .024 & 0.853 ± .023 & 0.657 ± .001 & 0.712 ± .002 & 0.654 ± .007 & 0.705 ± .007 & 0.594 ± .008 & 0.668 ± .006 & 0.536 ± .014 & 0.681 ± .011 \\
\hspace{0.0em} LLaMA 3.3 70B & 0.908 ± .002 & 0.903 ± .005 & 0.663 ± .004 & 0.731 ± .002 & 0.735 ± .003 & 0.800 ± .003 & 0.524 ± .002 & 0.624 ± .002 & 0.590 ± .005 & 0.732 ± .005 \\
\hspace{0.0em} Qwen 2.5 72B & 0.958 ± .002 & 0.978 ± .001 & 0.809 ± .003 & 0.856 ± .003 & 0.749 ± .003 & 0.808 ± .003 & 0.707 ± .001 & 0.785 ± .002 & 0.638 ± .001 & 0.745 ± .001 \\

\bottomrule
\end{tabular}
\label{tab:highlight_prompting_exp}
\end{table*}

The results are shown in Tables \ref{tab:prompting_exp0}, \ref{tab:retrieval_prompting_exp_tab}, and \ref{tab:highlight_prompting_exp}.

\subsection{Prompts Provided to LLMs}
\label{sec:prompts}
The system and user prompts used for the main SEER tasks are shown in Table \ref{tab:system_prompts} and \ref{tab:SEER_user_prompts}. The system prompts and user prompts used for the prompting experiments are shown in Tables \ref{tab:system_prompts} and \ref{tab:user_prompts}. 

\begin{table*}[t]
\centering
\small
\caption{User prompts for SEER Tasks, including Base and CoT variants.}
\begin{tabular}{p{0.95\textwidth}}
\toprule

\multicolumn{1}{c}{\textbf{User Base Prompt (Retrieve)}} \\
\midrule
*Begin Instructions* \\
You are given text. Some spans are emotionally expressive. \\
Return only the full unmodified emotionally expressive spans, and nothing else. \\
*End Instructions* \\
\texttt{\{text\}} \\

\midrule
\multicolumn{1}{c}{\textbf{User CoT Prompt (Retrieve)}} \\
\midrule
*Begin Instructions* \\
You are given text. Some spans are emotionally expressive. \\
Return the full unmodified emotionally expressive spans. \\

Reason step-by-step and explore the emotion content. Output "Reasoning:" and then your reasoning steps.
After reasoning, output "Response:" followed by the spans, and nothing else. \\
*End Instructions* \\
\texttt{\{text\}} \\

\midrule
\multicolumn{1}{c}{\textbf{User Prompt (Highlight)}} \\
\midrule
*Begin Instructions* \\
You are given text. Some spans are emotionally expressive. Surround the emotionally expressive spans with `**`. \\
Return only the full unmodified text with those markers, and nothing else. \\
*End Instructions* \\
\texttt{\{text\}} \\

\midrule
\multicolumn{1}{c}{\textbf{User CoT Prompt (Highlight)}} \\
\midrule
*Begin Instructions* \\
You are given text. Some spans are emotionally expressive. Surround the emotionally expressive spans with `**`. \\
Return the full unmodified text with those markers. \\

Reason step-by-step and explore the emotion content. Output "Reasoning:" and then your reasoning steps.
After reasoning, output "Response:" followed by the full unmodified text with those markers, and nothing else. \\
*End Instructions* \\
\texttt{\{text\}} \\

\bottomrule
\end{tabular}
\label{tab:SEER_user_prompts}
\end{table*}

\begin{table*}[t]
\centering
\small
\caption{System prompts used in both prompting experiments and main SEER tasks. The prompts for experiments 2 and 3 are used in the main SEER tasks.}
\begin{tabular}{p{0.95\textwidth}}
\toprule

\multicolumn{1}{c}{\textbf{System Prompt (Experiment 0)}} \\
\midrule
\textbf{Instructions} \\
1. \textbf{Immutable Input} \\
\hspace{1em}- Never delete, normalize, split, merge, or alter any original character (letters, apostrophes, punctuation, whitespace). \\[0.5ex]
2. \textbf{Self-Check} \\
\hspace{1em}- Verify that the remaining text is identical to the input. Retry until it passes. \\

\midrule
\multicolumn{1}{c}{\textbf{System Prompt (Experiment 2 and 3 Retrieval)}} \\
\midrule
\textbf{Instructions} \\
1. \textbf{Immutable Input} \\
\hspace{1em}- Never delete, normalize, split, merge, or alter any original character (letters, apostrophes, punctuation, whitespace). \\[0.5ex]
2. \textbf{Subjective Emotion Only} \\
\hspace{1em}- Only retrieve spans that reveal the speaker's internal emotional state or attitude. \\
\hspace{1em}- This includes: \\
\hspace{2em}- Explicit emotion words \\
\hspace{2em}- Implicit cues/phrases of feeling or reaction \\
\hspace{1em}- Do not retrieve: \\
\hspace{2em}- Purely factual or descriptive statements \\
\hspace{2em}- Neutral descriptions of events without any sentiment \\[0.5ex]
3. \textbf{Self-Check} \\
\hspace{1em}- Verify that the remaining text is identical to the input. Retry until it passes. \\
4. \textbf{Output} \\
\hspace{1em}- Return all spans on a single line, with each span separated by " | ". If there is only one span, do not include the " | ". \\
\hspace{1.6em}No headers, no metadata, no removed, added, or modified words.\\

\midrule
\multicolumn{1}{c}{\textbf{System Prompt (Experiment 1 Highlight)}} \\
\midrule
\textbf{Instructions} \\
1. \textbf{Immutable Input} \\
\hspace{1em}- You may only insert \texttt{**} markers. \\
\hspace{1em}- Never delete, normalize, split, merge, or alter any original character (letters, apostrophes, punctuation, whitespace). \\[0.5ex]
2. \textbf{Self-Check} \\
\hspace{1em}- After inserting your markers, remove all \texttt{**} and verify that the remaining text is identical to the input. \\ 
\hspace{1.6em}Retry until it passes. \\
\hspace{1em}- Any pair of \texttt{**} must surround the entire span. \\

\midrule
\multicolumn{1}{c}{\textbf{System Prompt (Experiment 2 and 3 Highlight)}} \\
\midrule
\textbf{Instructions} \\
1. \textbf{Immutable Input} \\
\hspace{1em}- You may only insert \texttt{**} markers. \\
\hspace{1em}- Never delete, normalize, split, merge, or alter any original character (letters, apostrophes, punctuation, whitespace). \\[0.5ex]
2. \textbf{Subjective Emotion Only} \\
\hspace{1em}- Only highlight spans that reveal the speaker's internal emotional state or attitude. \\
\hspace{1em}- This includes: \\
\hspace{2em}- Explicit emotion words \\
\hspace{2em}- Implicit cues/phrases of feeling or reaction \\
\hspace{1em}- Do not mark: \\
\hspace{2em}- Purely factual or descriptive statements \\
\hspace{2em}- Neutral descriptions of events without any sentiment \\[0.5ex]
3. \textbf{Self-Check} \\
\hspace{1em}- After inserting your markers, remove all \texttt{**} and verify that the remaining text is identical to the input. Retry until it passes. \\
\hspace{1em}- If the input is completely neutral, return it unchanged, with no markers. \\
\hspace{1em}- Any pair of \texttt{**} must surround the entire span. \\[0.5ex]
4. \textbf{Output} \\
\hspace{1em}- Return only the marked text. No headers, no metadata, no removed, added, or modified words. \\
\bottomrule
\end{tabular}
\label{tab:system_prompts}
\end{table*}

\begin{table*}[t]
\centering
\footnotesize
\caption{User prompts used across experiments. Placeholders \texttt{\{text\}} and \texttt{\{target\_sentence\}} represent inputs given during evaluation.}
\begin{tabular}{p{0.95\textwidth}}
\toprule
\multicolumn{1}{c}{\textbf{User Prompt (Experiment 0)}} \\
\midrule
*Begin Instructions* \\
You are given a series of sentences. Return only the full unmodified text, and nothing else. \\
*End Instructions* \\
\texttt{\{text\}} \\
\midrule
\multicolumn{1}{c}{\textbf{User Prompt (Experiment 1: Highlight)}} \\
\midrule
*Begin Instructions* \\
You are given a series of sentences, which contains target sentence: \texttt{\{target\_sentence\}}. Surround the target sentence with \texttt{**}: \\
Return only the full unmodified text with those markers, and nothing else. \\
*End Instructions* \\
\texttt{\{text\}} \\

\midrule
\multicolumn{1}{c}{\textbf{User Prompt (Experiment 1: Retrieval)}} \\
\midrule
*Begin Instructions* \\
You are given a series of sentences, which contains target sentence: \texttt{\{target\_sentence\}}. \\
Return only the full unmodified target sentence, and nothing else. \\
*End Instructions* \\
\texttt{\{text\}} \\

\midrule
\multicolumn{1}{c}{\textbf{User Prompt (Experiment 2: Highlight)}} \\
\midrule
*Begin Instructions* \\
You are given a series of sentences. One sentence is emotionally expressive. Surround the emotionally expressive sentence with \texttt{**}: \\
Return only the full unmodified text with those markers, and nothing else. \\
*End Instructions* \\
\texttt{\{text\}} \\

\midrule
\multicolumn{1}{c}{\textbf{User Prompt (Experiment 2: Retrieval)}} \\
\midrule
*Begin Instructions* \\
You are given a series of sentences. One sentence is emotionally expressive. \\
Return only the full unmodified emotionally expressive sentence, and nothing else. \\
*End Instructions* \\
\texttt{\{text\}} \\
\bottomrule
\end{tabular}
\label{tab:user_prompts}
\end{table*}

For GPT annotation, we follow the approach of \citet{niu2024text} and provide the instructions and labeling schema in the system prompt, and provide the transcript itself in the user prompt.

The GPT annotation system prompt is: \textit{"You are an emotionally-intelligent and empathetic agent. You will be given a piece of text, and your task is to identify the emotions expressed by the speaker. You are only allowed to make one selection from the following emotions: \{set of emotions\}. Do not return anything else."}

\section{Task 2 Emotion Transition Distribution}
\label{sec:task2_emotion_transition_distribution}
\begin{figure*}[t]
    \centering
    \begin{subfigure}[t]{0.45\textwidth}
        \centering
        \includegraphics[width=\textwidth]{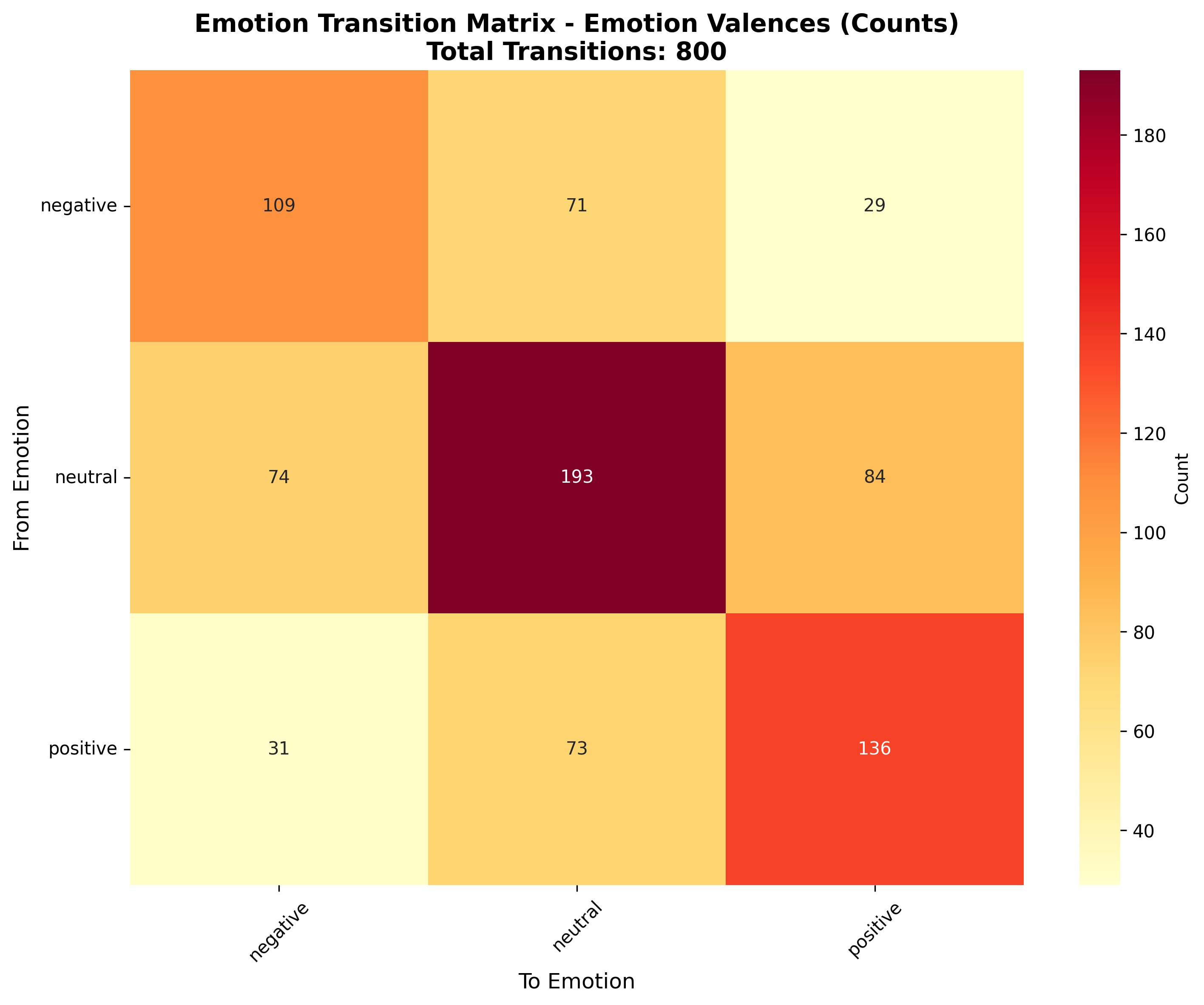}
        \caption{Valence transitions.}
        \label{fig:task2_valence_transitions}
    \end{subfigure}
    \hfill
    \begin{subfigure}[t]{0.45\textwidth}
        \centering
        \includegraphics[height=6cm]{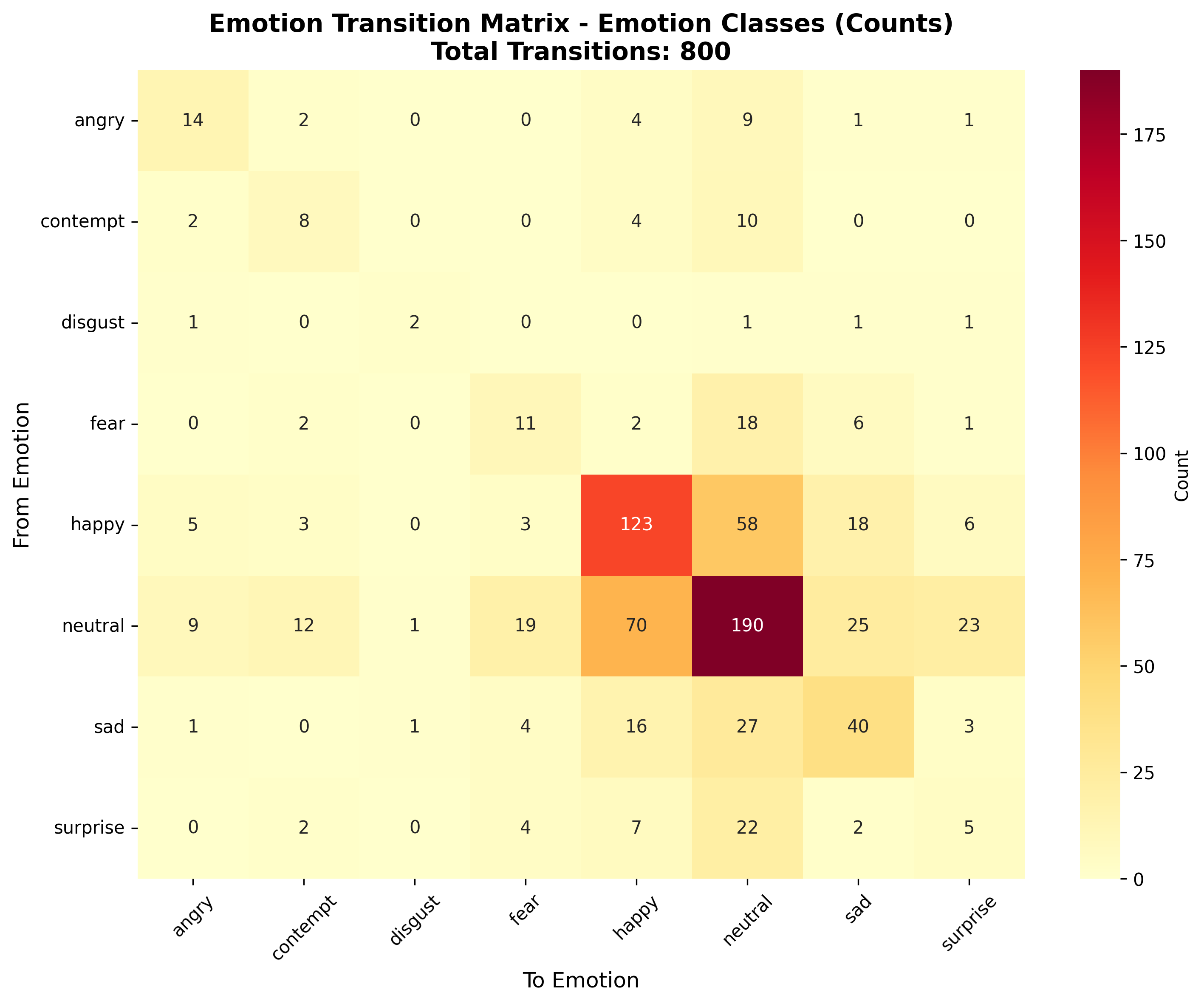}
        \caption{Categorical emotion transitions.}
        \label{fig:task2_categorical_transitions}
    \end{subfigure}
    \caption{Emotion transitions between adjacent sentences for Task 2.}
    \label{fig:task2_emotion_transitions}
\end{figure*}

Figure \ref{fig:task2_emotion_transitions} describes the emotion transitions between adjacent sentences for Task 2. The most common transition is neutral to neutral. Figure \ref{fig:task2_valence_transitions} shows the valence transitions, and Figure \ref{fig:task2_categorical_transitions} shows the categorical emotion transitions.

\section{Annotator Recruitment}
\label{sec:annotator_recruitment}
We recruited annotators using the Prolific platform\footnote{prolific.com}. We separated the 200 samples per task into surveys with 20 samples each to reduce annotation fatigue. We recruited three annotators per survey and paid them at a rate of \$10 an hour. We recruited annotators that were (1) native English speakers, (2) residents of the USA.

The instructions were as follows: "In the following task, you will be asked to identify emotionally-relevant text. You will be presented with short passages and asked to identify the emotional text within the passage. Please use your mouse to highlight the regions of text that are emotionally salient. Ensure that you highlight only the specific text that expresses emotion." The participants were then asked to consent to the following: "(1) I have read and understood the information above, (2) I understand I might see potentially offensive or sexual content, and (3) I want to participate in this research and continue with the study," before proceeding to the main task.

\section{Emotion Category Errors}
\label{sec:emotion_category_errors}

\begin{figure*}[t]
    \centering
    \begin{subfigure}[t]{0.45\textwidth}
        \centering
        \includegraphics[width=\textwidth]{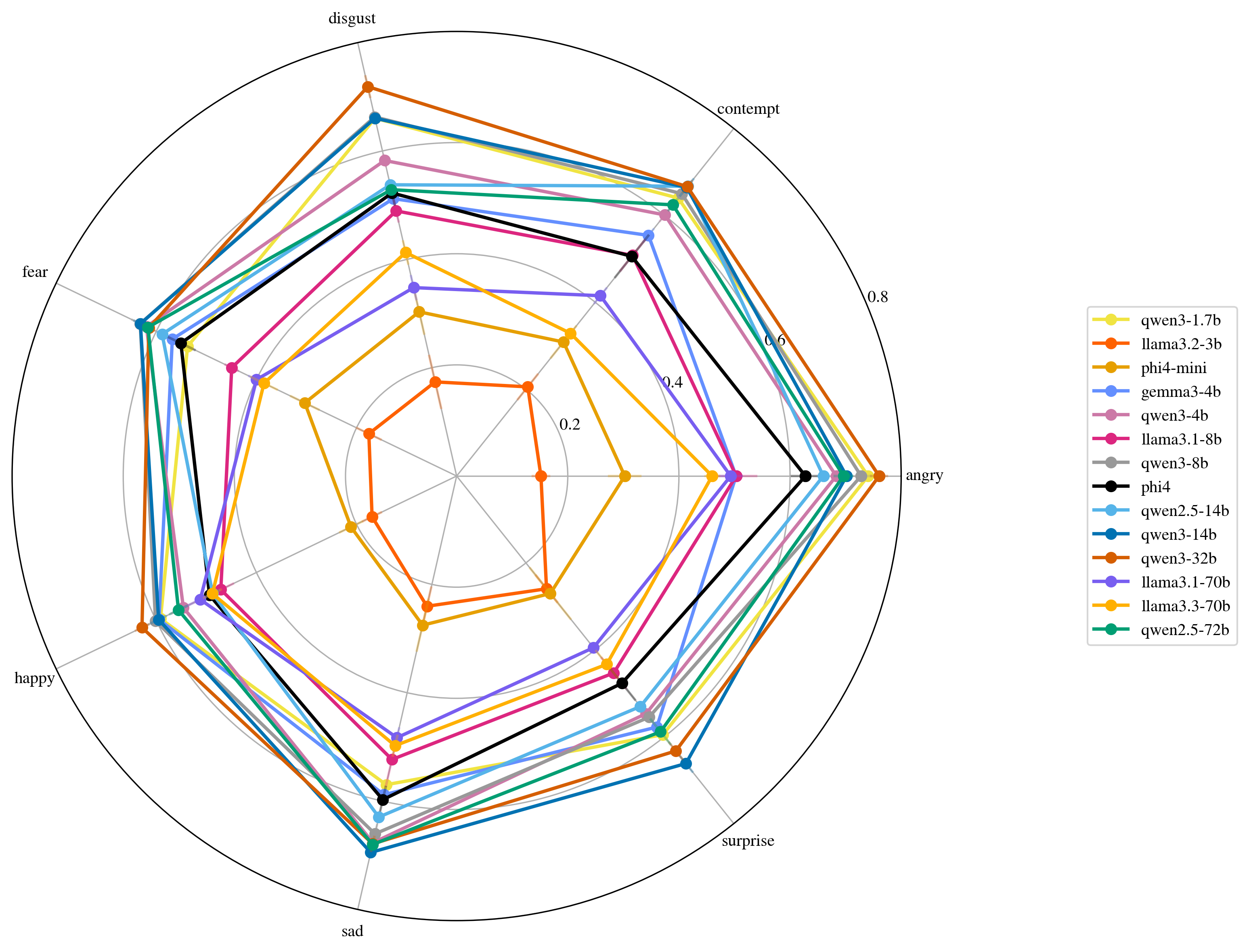}
        \caption{Emotion-wise F1 scores for Task 1 (Retrieve-Base).}
        \label{fig:task1_retrieve_categorical}
    \end{subfigure}
    \hfill
    \begin{subfigure}[t]{0.45\textwidth}
        \centering
        \includegraphics[height=5.5cm]{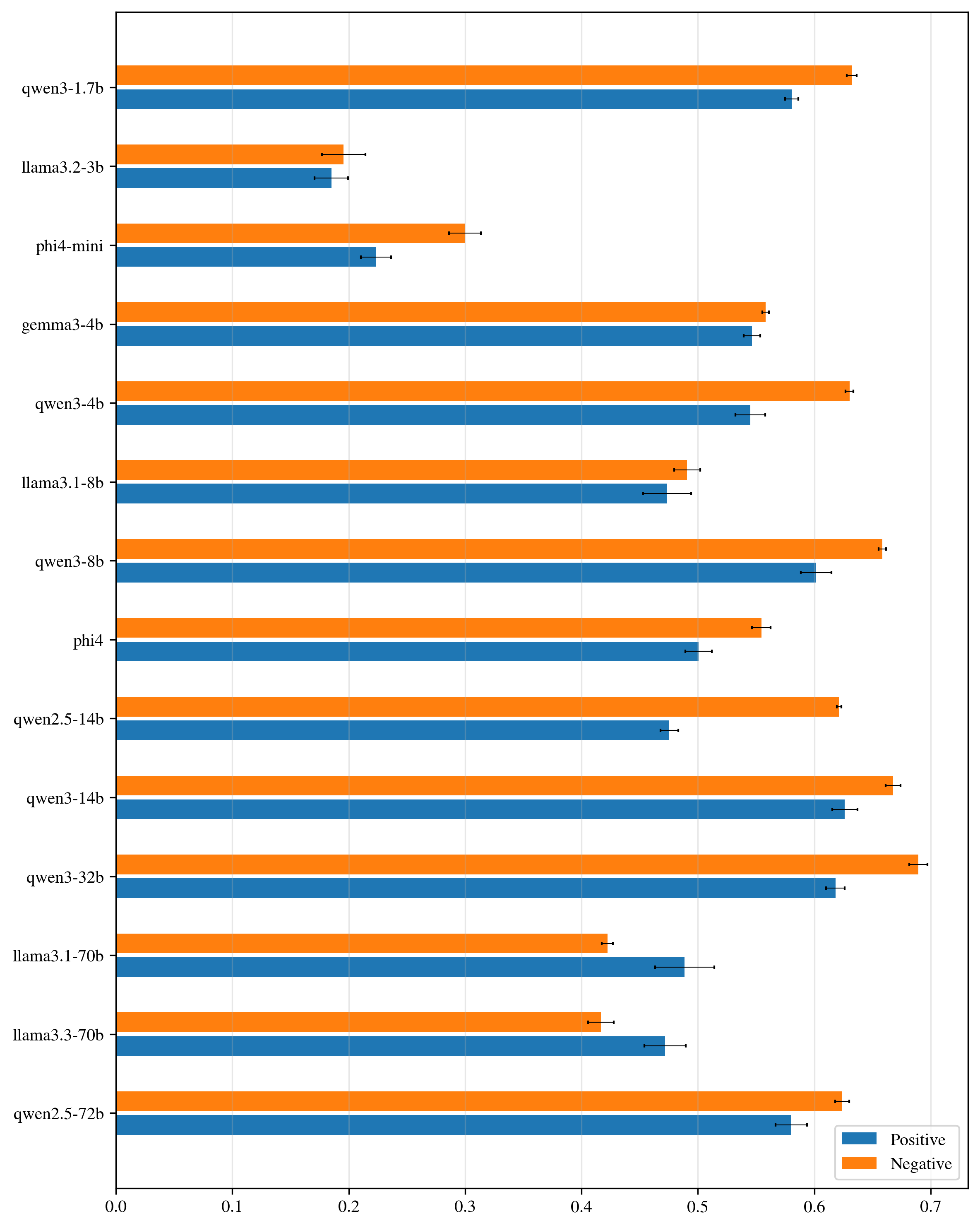}
        \caption{Valence F1 scores for Task 1 (Retrieve-Base).}
        \label{fig:task1_retrieve_valence}
    \end{subfigure}
    \caption{Task 1 (Retrieve-Base). (a) Per-emotion F1 scores. (b) Per-valence F1 scores.}
    \label{fig:task1_retrieve_errors}
\end{figure*}

\begin{figure*}[t]
    \centering
    \begin{subfigure}[t]{0.45\textwidth}
        \centering
        \includegraphics[width=\textwidth]{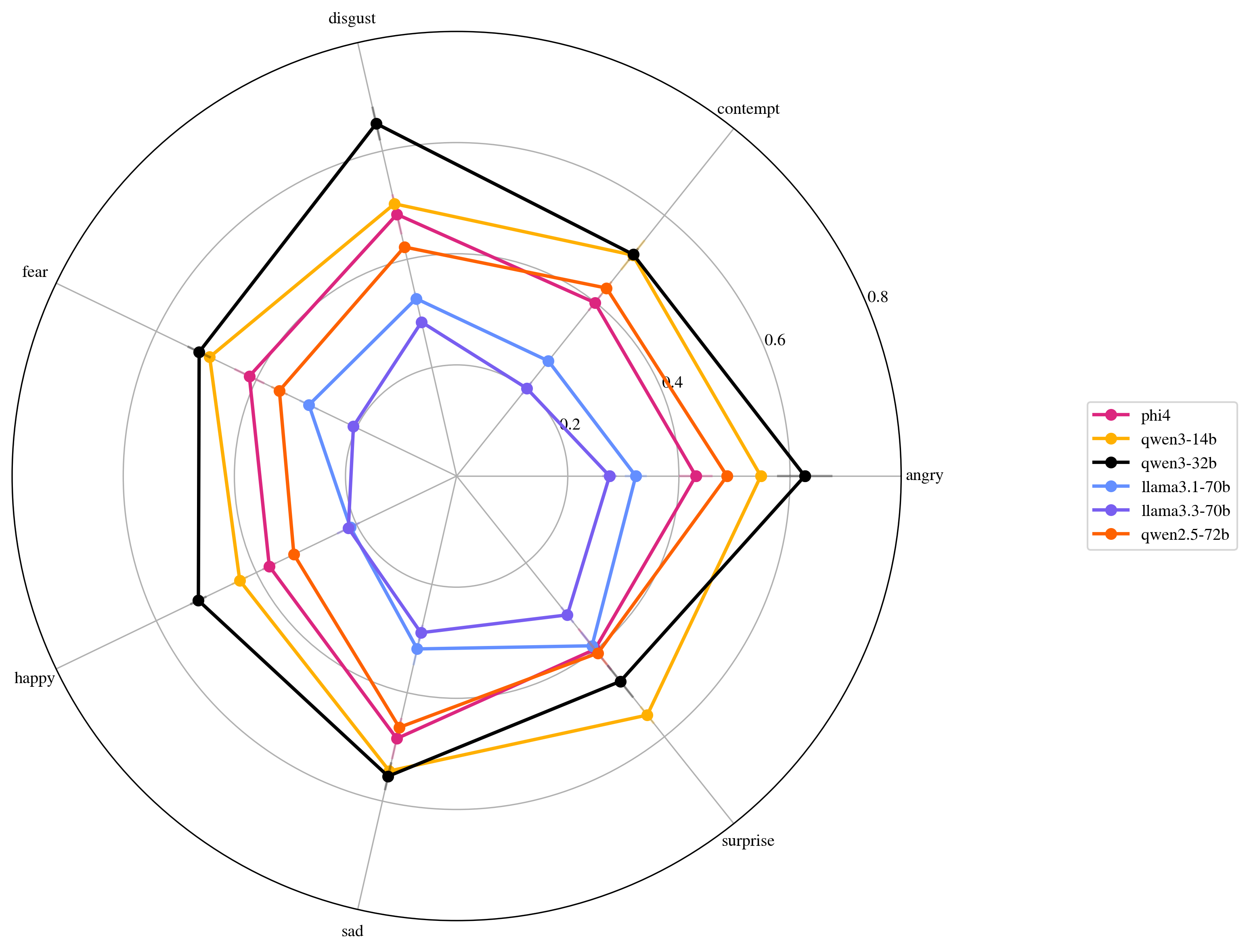}
        \caption{Emotion-wise F1 scores for Task 1 (Highlight-Base).}
        \label{fig:task1_highlight_categorical}
    \end{subfigure}
    \hfill
    \begin{subfigure}[t]{0.45\textwidth}
        \centering
        \includegraphics[height=5.5cm]{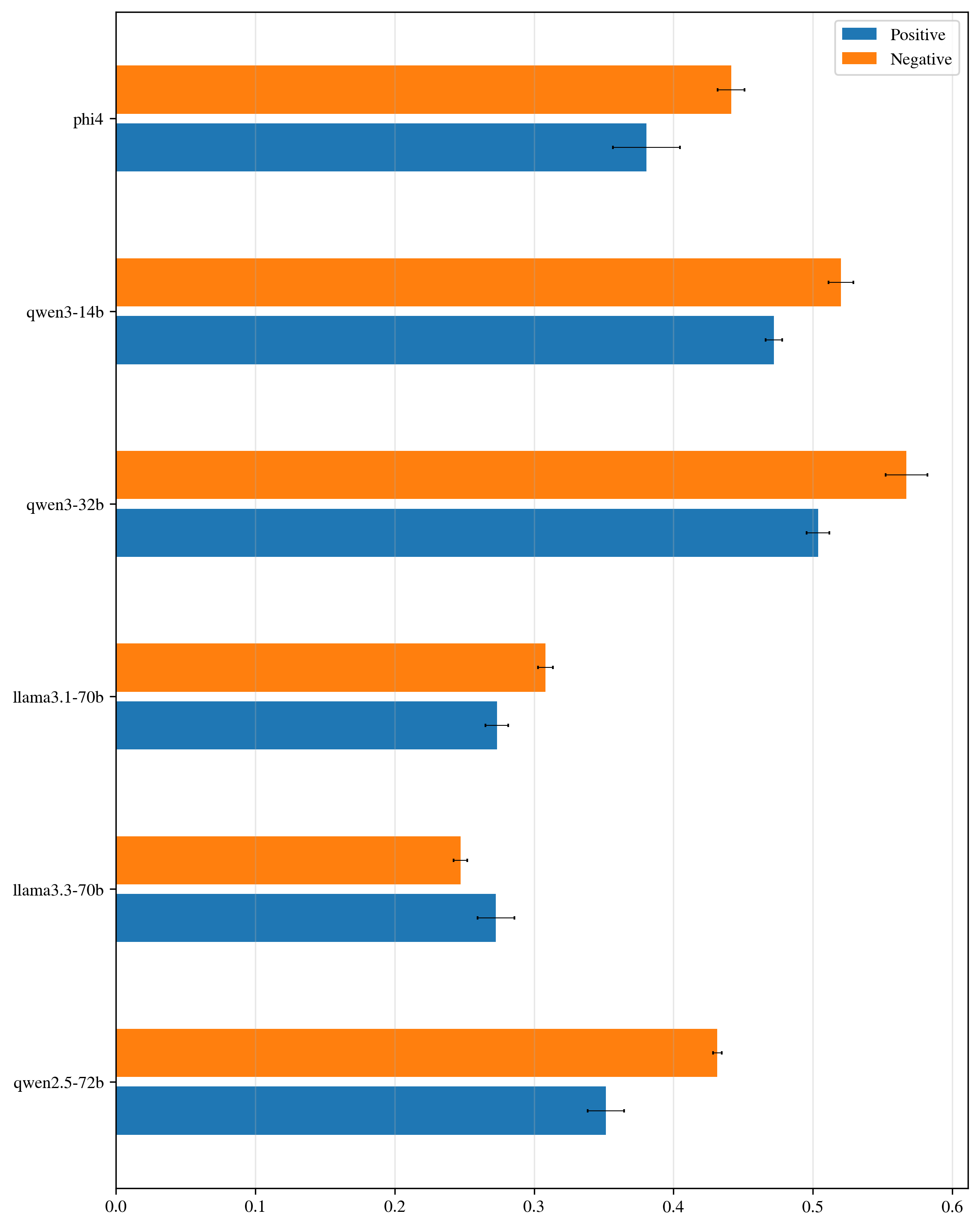}
        \caption{Valence F1 scores for Task 1 (Highlight-Base).}
        \label{fig:task1_highlight_valence}
    \end{subfigure}
    \caption{Task 1 (Highlight-Base). (a) Per-emotion F1 scores. (b) Per-valence F1 scores.}
    \label{fig:task1_highlight_errors}
\end{figure*}

\begin{figure*}[t]
    \centering
    \begin{subfigure}[t]{0.45\textwidth}
        \centering
        \includegraphics[width=\textwidth]{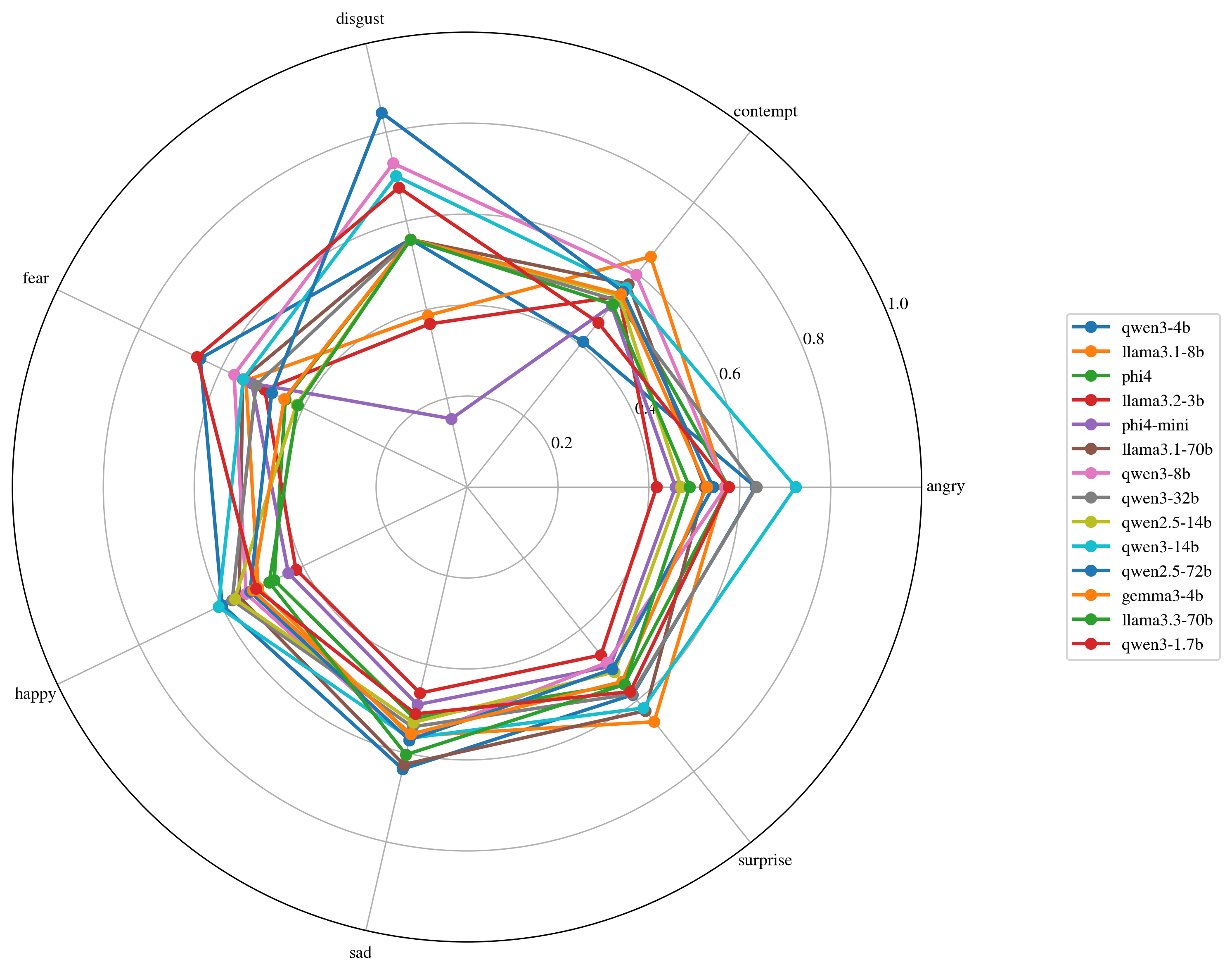}
        \caption{Categorical emotion F1 (not including neutral sentences).}
        \label{fig:task2_retrieve_categorical_errors}
    \end{subfigure}
    \hfill
    \begin{subfigure}[t]{0.45\textwidth}
        \centering
        \includegraphics[height=5cm]{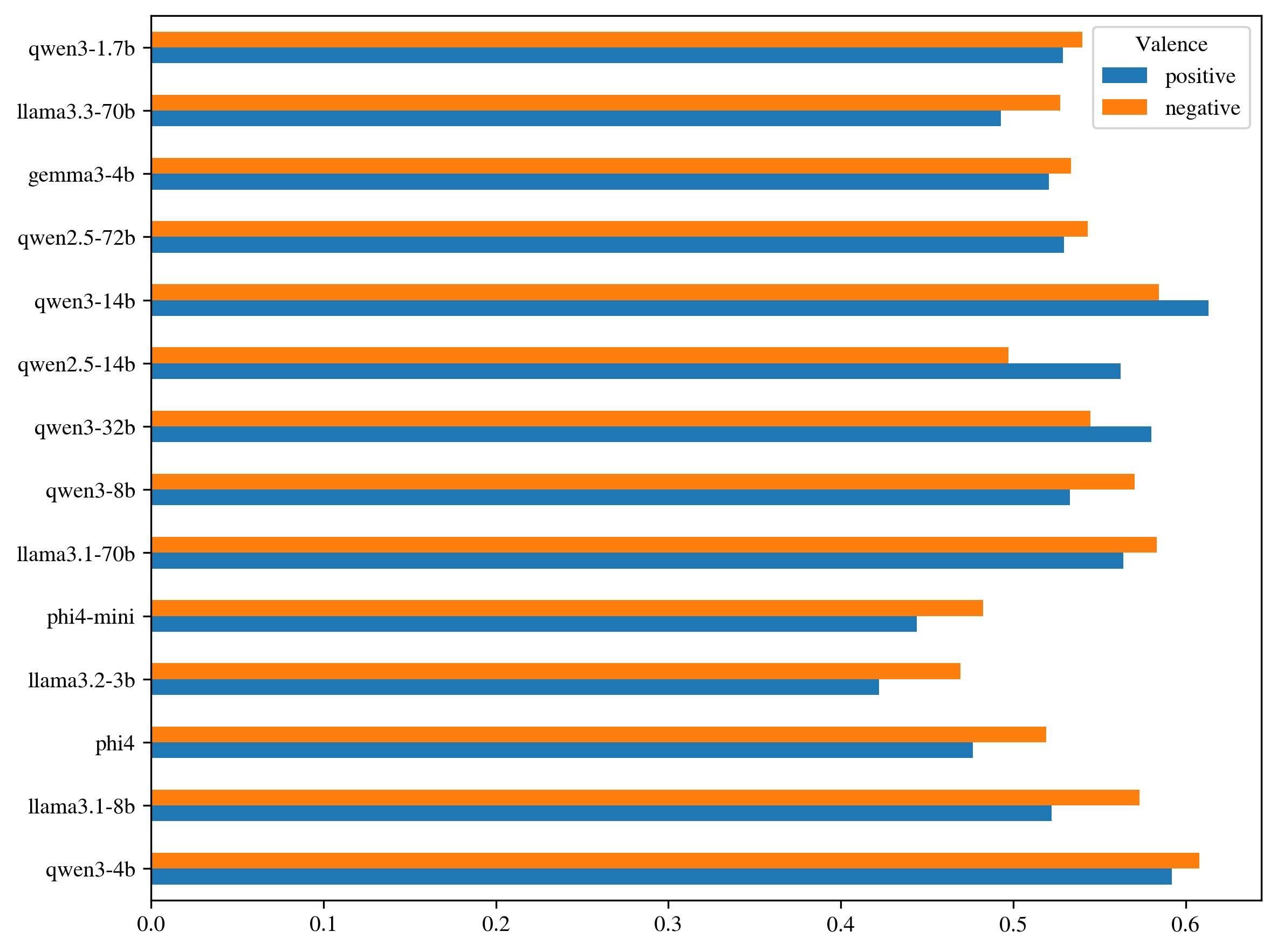}
        \caption{Valence F1 (not including neutral sentences).}
        \label{fig:task2_retrieve_valence_errors}
    \end{subfigure}
    
    \vspace{1em}
    
    \begin{subfigure}[t]{0.7\textwidth}
        \centering
        \includegraphics[width=\textwidth]{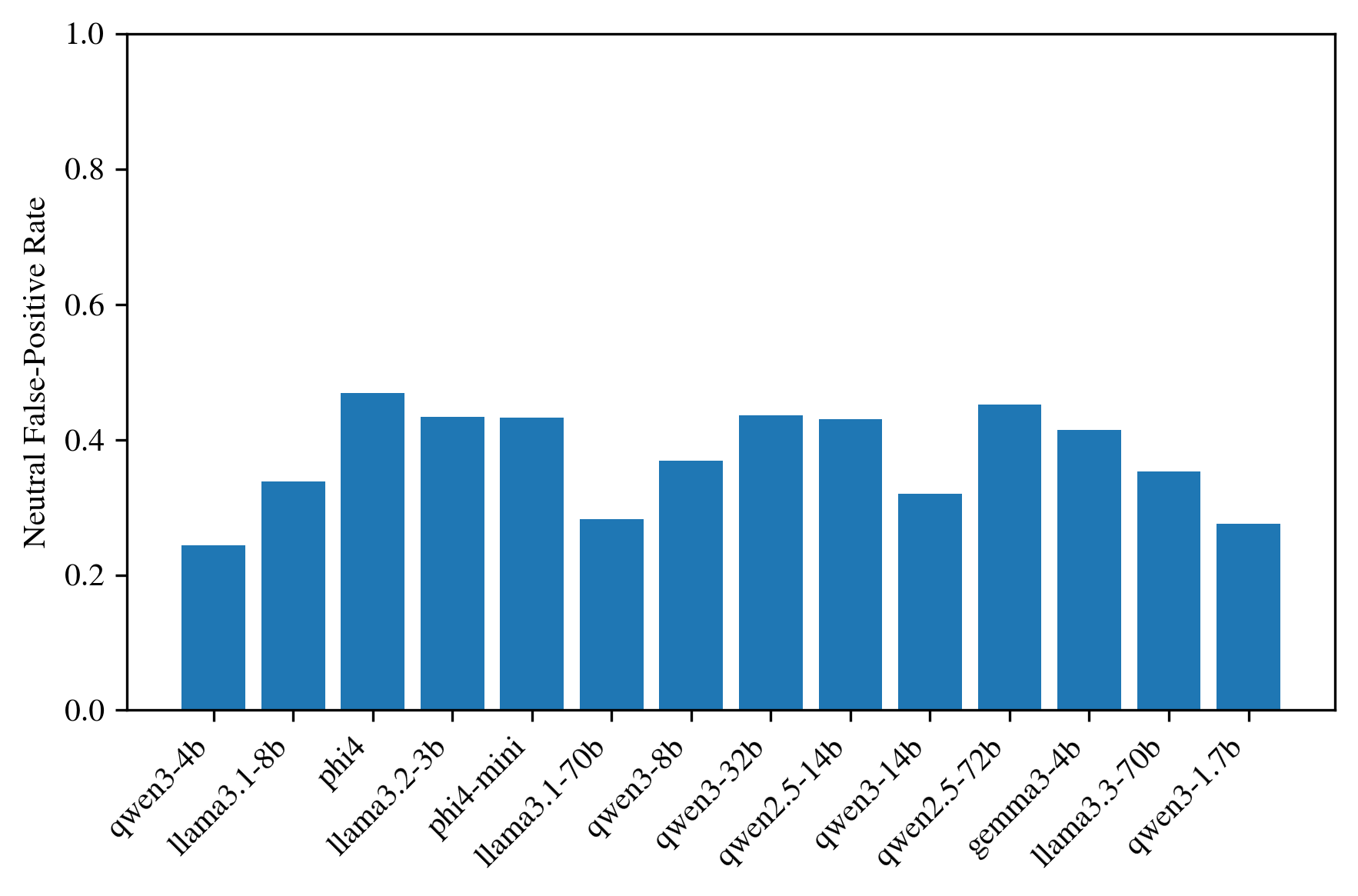}
        \caption{Neutral False Positive Rate (NFPR): proportion of predicted spans in gold-labeled neutral sentences.}
        \label{fig:task2_retrieve_nfpr}
    \end{subfigure}
    \caption{Emotion category errors in Task 2 Retrieve-Base.}
    \label{fig:task2_retrieve_base_emotion_category_errors}
\end{figure*}

\begin{figure*}[t]
    \centering
    \begin{subfigure}[t]{0.45\textwidth}
        \centering
        \includegraphics[width=\textwidth]{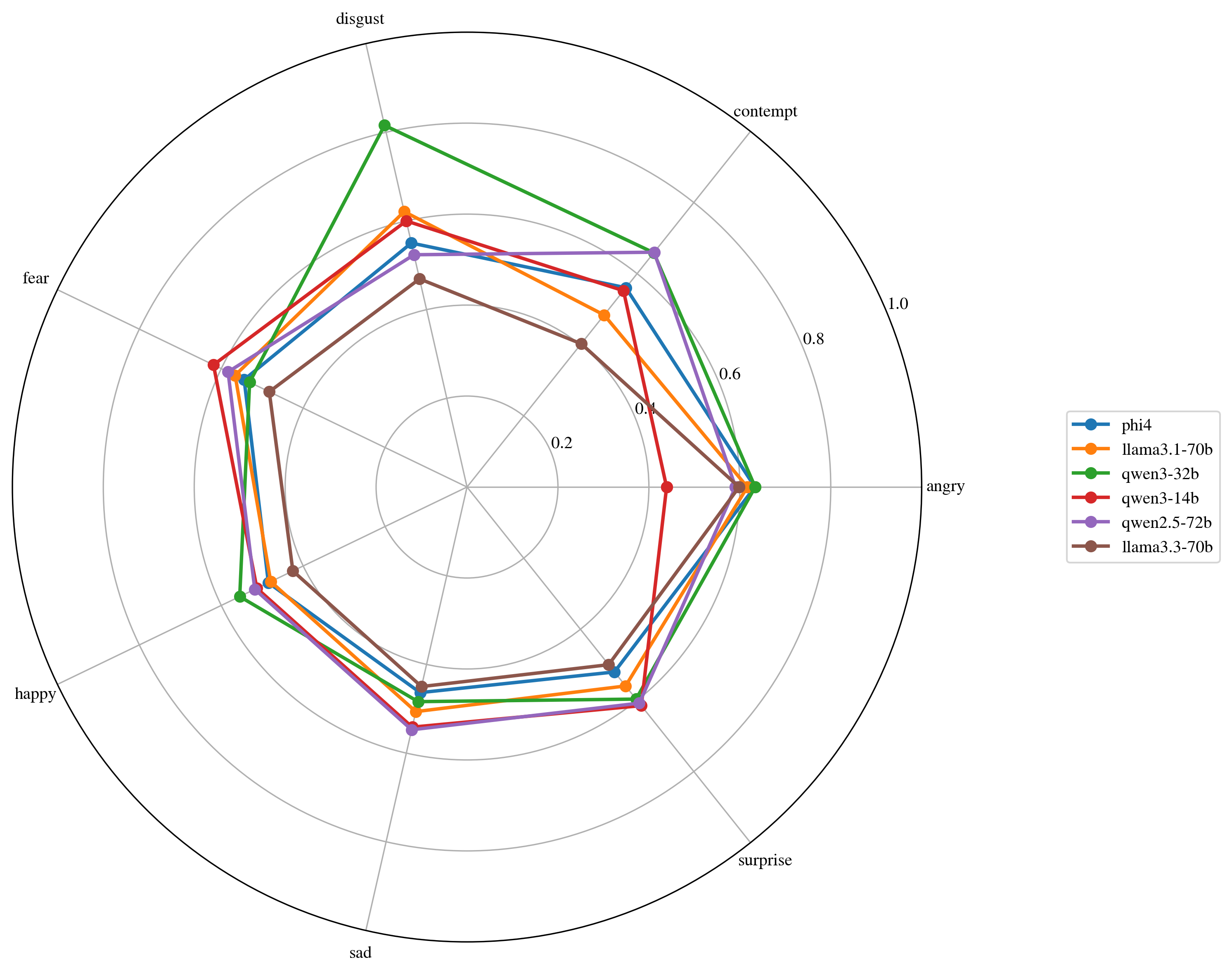}
        \caption{Categorical emotion F1 (not including neutral sentences).}
        \label{fig:task2_highlight_categorical_errors}
    \end{subfigure}
    \hfill
    \begin{subfigure}[t]{0.45\textwidth}
        \centering
        \includegraphics[height=6cm]{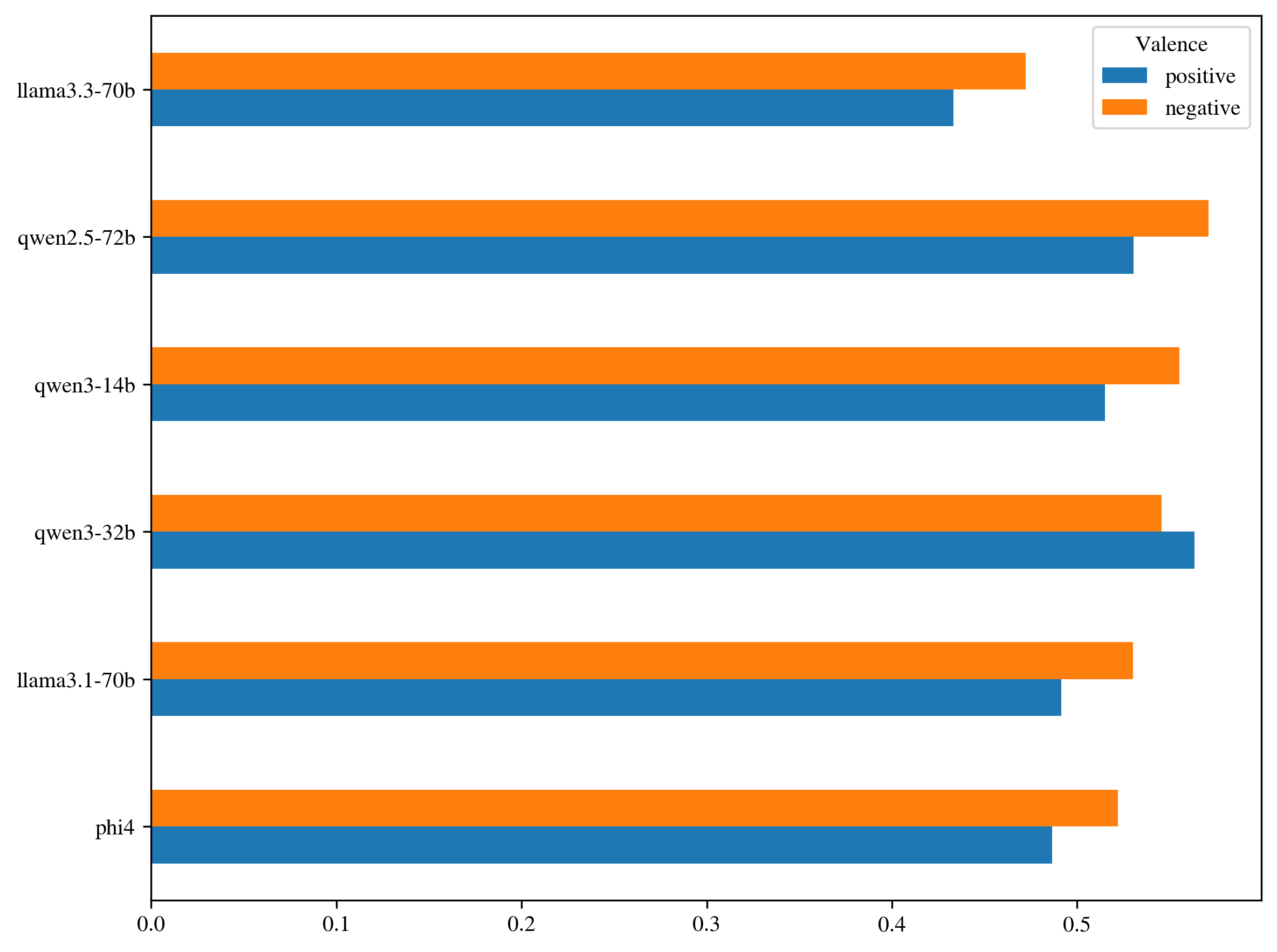}
        \caption{Valence F1 (not including neutral sentences).}
        \label{fig:task2_highlight_valence_errors}
    \end{subfigure}
    
    \vspace{1em}
    
    \begin{subfigure}[t]{0.7\textwidth}
        \centering
        \includegraphics[width=\textwidth]{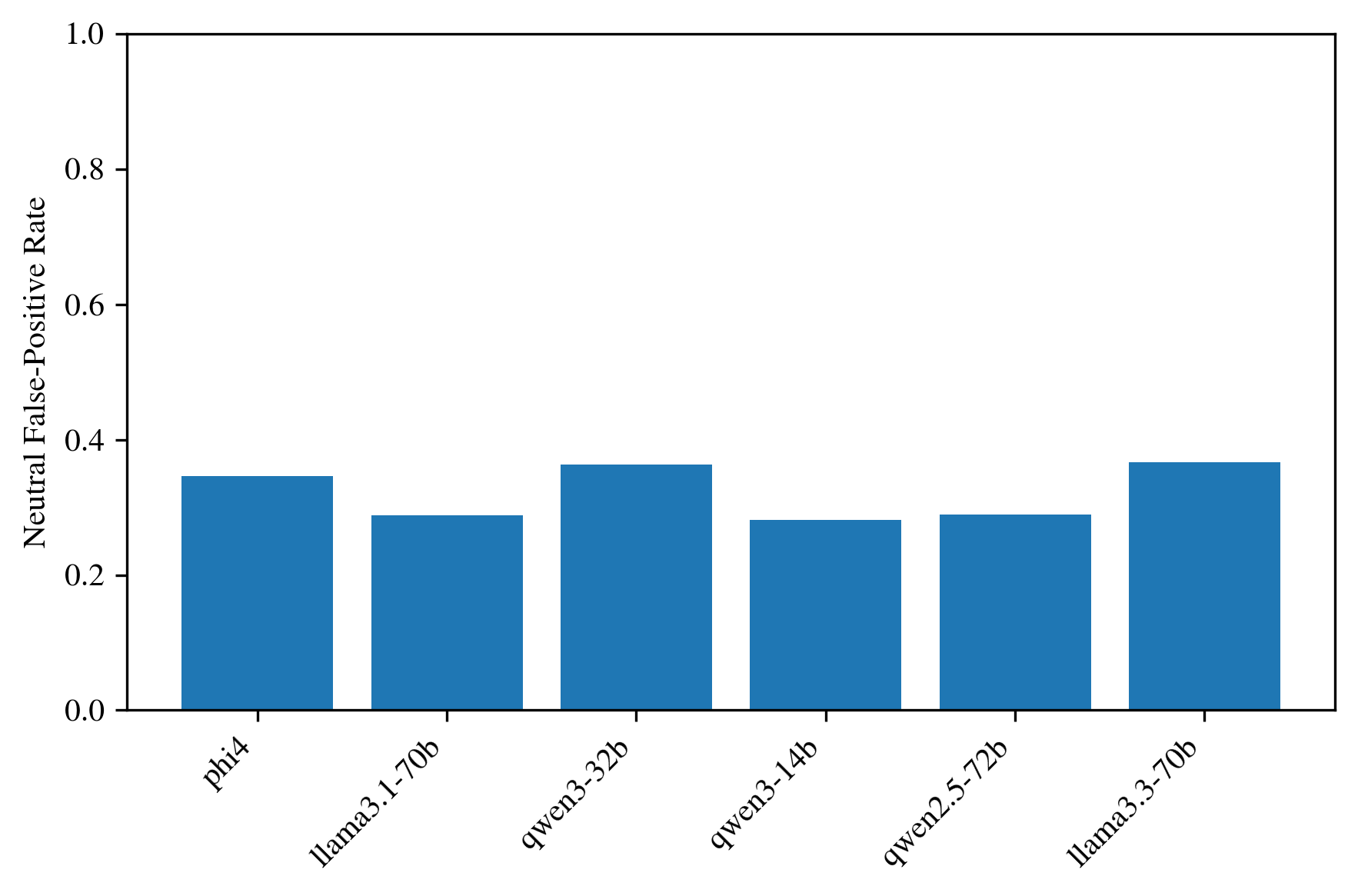}
        \caption{Neutral False Positive Rate (NFPR): proportion of predicted spans in gold-labeled neutral sentences.}
        \label{fig:task2_highlight_nfpr}
    \end{subfigure}
    \caption{Emotion category errors in Task 2 Highlight-Base.}
    \label{fig:task2_highlight_base_emotion_category_errors}
\end{figure*}

See Figures \ref{fig:task1_retrieve_errors} and \ref{fig:task1_highlight_errors} for Task 1, \ref{fig:task2_retrieve_base_emotion_category_errors} and \ref{fig:task2_highlight_base_emotion_category_errors} for Task 2.

\section{Model Checkpoints}
\begin{table*}[t]
\centering
\small
\caption{Hugging Face model checkpoint names.}
\begin{tabular}{l c r}
\toprule
\textbf{Model} & \textbf{Reasoning} & \textbf{Hugging Face Checkpoint} \\
\midrule
Qwen 3 0.6B & \checkmark & Qwen/Qwen3-0.6B \\
LLaMA 3.2 1B & $\times$ & meta-llama/Llama-3.2-1B-Instruct \\
Qwen 3 1.7B & \checkmark & Qwen/Qwen3-1.7B \\
LLaMA 3.2 3B & $\times$ & meta-llama/Llama-3.2-3B-Instruct \\
Phi 4 Mini 3.8B & $\times$ &  microsoft/Phi-4-mini-instruct \\
Qwen 3 4B & \checkmark &  Qwen/Qwen3-4B \\
Gemma 3 4B & $\times$ &  google/gemma-3-4b-it \\
LLaMA 3.1 8B & $\times$ &  meta-llama/Llama-3.1-8B-Instruct \\
Qwen 3 8B & \checkmark & Qwen/Qwen3-8B \\
Phi 4 14B & $\times$ &  microsoft/phi-4 \\
Qwen 2.5 14B & $\times$ & Qwen/Qwen2.5-14B-Instruct \\
Qwen 3 14B & \checkmark & Qwen/Qwen3-14B \\
Qwen 3 32B & \checkmark & Qwen/Qwen3-32B \\
LLaMA 3.1 70B & $\times$ &  meta-llama/Llama-3.1-70B-Instruct \\
LLaMA 3.3 70B & $\times$ &  meta-llama/Llama-3.3-70B-Instruct \\
Qwen 2.5 72B & $\times$ &  Qwen/Qwen2.5-72B-Instruct \\
\bottomrule
\end{tabular}
\label{tab:hf_checkpoints}
\end{table*}

The exact model checkpoints from huggingface transformers library are in Table \ref{tab:hf_checkpoints}.

\end{document}